%% file: main.tex
\documentclass{article}

    \PassOptionsToPackage{numbers, compress}{natbib}

\usepackage[preprint]{neurips_2026}

\usepackage[utf8]{inputenc} %
\usepackage[T1]{fontenc}    %
\usepackage[table]{xcolor}
\usepackage{hyperref}       %
\usepackage{url}            %
\usepackage{booktabs}       %
\usepackage{amsmath, amsfonts, amssymb}       %
\usepackage{nicefrac}       %
\usepackage{microtype}      %
\usepackage{adjustbox}
\usepackage{multirow}
\usepackage{xspace}
\usepackage{cleveref}
\usepackage{subcaption}
\usepackage{listings}
\usepackage{enumitem}
\usepackage{fancyvrb}
\usepackage{fvextra}

\usepackage{tikz}
\usepackage{fontawesome5}
\usetikzlibrary{positioning, fit, shadows, calc, backgrounds}
\usepackage{graphicx}
\usepackage{tcolorbox}

\definecolor{coralhl}{RGB}{216,90,48}

\definecolor{tealhl}{RGB}{15,110,75}
\definecolor{tealtext}{RGB}{8,80,55}

\newcommand{\teal}[1]{%
  \colorbox{tealhl!20}{%
    \textcolor{tealtext}{ #1}%
  }%
}

\definecolor{promptorange}{RGB}{230,132,0}
\definecolor{abstgreen}{RGB}{96,160,74}
\definecolor{badred}{RGB}{220,0,0}
\definecolor{promptbg}{RGB}{235,235,235}
\definecolor{goodbg}{RGB}{213,231,202}
\definecolor{badbg}{RGB}{243,210,210}
\definecolor{icongray}{RGB}{95,95,95}
\definecolor{charcoal}{rgb}{0.21, 0.27, 0.31}

\definecolor{headerbg}{RGB}{67,67,67}

\newtcolorbox{promptpanel}[1]{
  colback=gray!3!white,
  colframe=charcoal,
  title={#1},
  colbacktitle=charcoal,
  coltitle=white,
  fonttitle=\bfseries\large,
  fontupper=\small
}

\newcommand{\tool}{\textsc{AGOPS}\xspace}

\newcommand{\task}{\mathcal{T}}

\newcommand{\query}{\boldsymbol{q}}

\newcommand{\answer}{\boldsymbol{a}}
\newcommand{\user}{\mathcal{U}}
\newcommand{\writer}{\mathcal{W}}
\newcommand{\prompt}{\rho}
\newcommand{\LLM}{\mathtt{LLM}}

\title{Automatically Evolving Prompt Guidelines for Task-Specific Optimization}

\author{Cedric Richter$^{1}$, Salah Ghamizi$^{1,2}$, Mike Papadakis$^{1}$ \\
    $^{1}$SnT , University of Luxembourg, Luxembourg. first.last@uni.lu \\
    $^{2}$Luxembourg Institute of Health (LIH), Luxembourg
}

\begin{document}

\maketitle

\begin{abstract}
For Large Language Models to reliably answer user queries, users must clearly specify requirements, context, and constraints. In practice, however, user queries are often underspecified, forcing models to infer unstated assumptions that may misalign with the actual user intent. Existing prompt engineering guidelines aim to mitigate this issue, they are typically generic and task-agnostic, limiting their practical utility. Additionally, existing guidelines are formed manually and in a non-systematic way. To this  end, we study {\em prompt guideline optimization}: the problem of automatically generating {\em task-specific} guidelines that help write better-specified prompts for a given task and model. Our key observation is that existing (completed) task examples (aka reference answers) often implicitly encode the missing information required to complete underspecified queries, including behavioral constraints, contextual assumptions, and evaluation criteria. We therefore propose $\tool$, an automatic approach that evolves task-specific guidelines via an optimization scheme that involves a prompt LLM writer, a solver LLM and prompt evolution, which maximize downstream effectiveness on a set of examples (user queries with reference answers). At inference time, our guidelines help users write well-specified prompts, boosting the effectiveness of LLMs. We show across mathematical reasoning, medical question answering, and coding tasks, that prompt underspecification leads to major drops (up to 95.3\%) in downstream task performance (compared to well-specified prompts) and, perhaps more importantly, that this drop can hardly be recovered by existing prompt optimization techniques. Users following $\tool$ guidelines can regain this loss (increasing performance between 15.5 to 81.7\% on average) consistently across all benchmarks. %

\end{abstract}

\section{Introduction}
Large Language Models (LLMs) are applied in high-stakes domains such as medical diagnosis, software engineering, legal analysis, and financial decision making~\cite{DBLP:journals/tmlr/0002MZH0NYMPW24}. Unfortunately, even small errors in LLM outputs can have serious consequences~\cite{DBLP:journals/tmlr/0002MZH0NYMPW24, DBLP:journals/corr/abs-2401-01301, maity2025large}. To address this issue, users must write clear and complete requirements, assumptions, and constraints that reliably guide the LLM to the desired outcome~\cite{DBLP:conf/chi/Zamfirescu-Pereira23, DBLP:journals/corr/abs-2503-16789}. 
In practice, users are often unaware of what they know~\cite{DBLP:journals/ijinfoman/KucharskaE23}, what they assume~\cite{DBLP:journals/corr/abs-2008-08849}, and how to elicit their knowledge in a conversation with an LLM~\cite{DBLP:conf/chi/Zamfirescu-Pereira23, DBLP:journals/corr/abs-2503-16789}. As a consequence, user prompts are often {\em underspecified}~\cite{DBLP:journals/corr/abs-2505-13360}, lacking critical details. This is particularly evident in non-expert users, who tend to omit details that are necessary or important~\cite{DBLP:conf/chi/Zamfirescu-Pereira23, DBLP:conf/acl/BabeNZGFA24}. Our results confirm this issues and show that prompt underspecification leads to major drops (up to 95.3\%) in downstream task performance compared to well-specified prompts.

\begin{figure}[t]
\centering
\begin{subfigure}{0.32\linewidth}
    \resizebox{\linewidth}{!}{\input{figures/noguide}}
    \caption{Generic Guideline}\label{fig:noguide}
\end{subfigure}
\begin{subfigure}{0.32\linewidth}
     \resizebox{\linewidth}{!}{\input{figures/guide}}
    \caption{Task-Specific Guideline}\label{fig:guide}
\end{subfigure}
\begin{subfigure}{0.32\linewidth}
    \includegraphics[width=\linewidth]{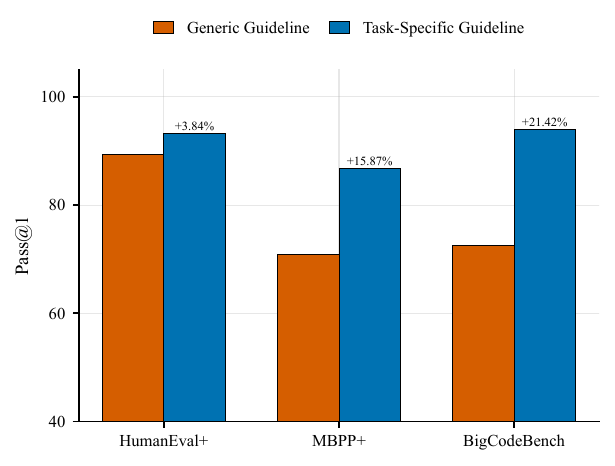}
    \caption{Generic vs. Task-Specific}\label{fig:code-intro}
\end{subfigure}
\caption{\textbf{Task-specific guidelines.}  (a) Users lack concrete direction on how to fully specify their prompts, e.g., omitting details such as order preservation of the output. (b) Task-specific guidelines provide concrete recommendations to the task (e.g. specifying concrete output constraints for coding problems) (c)  Task-specific guidelines lead to better specified prompts than generic ones.}
\label{fig:motivation}
\end{figure}

Prompt engineering guidelines aim to address this by helping the user to write {\em better-specified prompts}~\cite{DBLP:journals/corr/abs-2406-06608}, prescribing what context to include, what requirements to state explicitly, and which prompting strategies to apply such as chain-of-thought prompting~\cite{DBLP:conf/nips/Wei0SBIXCLZ22} or the use of assertive language~\cite{DBLP:journals/corr/abs-2601-13118}. However, these guidelines are often formulated manually, in a non-systematic way and are overly generic and task-agnostic. %
\Cref{fig:noguide} illustrates this point through a coding example: code generation guidelines~\cite{DBLP:journals/corr/abs-2601-13118} commonly instruct users to specify pre- and postconditions in their prompt, but a user unaware of which conditions matter for their task, such as order preservation when retrieving unique elements from a list, gains little from such generic advice. 

We introduce the problem of {\em prompt guideline optimization;} the problem of automatically synthesizing task-specific prompt engineering guidelines based on a set of example resolved tasks. By following these guidelines, users formulate well-specified prompts that boost the effectiveness of LLMs. Unlike existing guidelines that summarize general or domain-specific heuristics, our approach focuses on the generation of concrete, actionable instructions tailored to a specific task and model (see \Cref{fig:guide}). 

The key challenge involved in this problem is the identification and optimization of the sought {\em user knowledge} (assumptions, context, and output constraints) that must be specified by the user when writing a prompt for a target task. This knowledge is typically known by the user, but is absent in the user-defined prompts since the user is unaware of its importance. It is noted that such omissions (underspecification) are often critical in guiding the LLM to perform the intended task. 

Our key idea is to rely on resolved tasks to extract relevant information about the key elements that the task-specific prompts need to include. For instance, we aim at extracting implicitly encoded task-specific user knowledge from reasoning traces or a reference implementations to evolve guidelines. At the same time, to ensure generalization to unseen tasks/case it is imperative that the guidelines are to some degree general, meaning that they should not include solution-specific elements. To this end we aim at explicitly constraining any copying from the resolved tasks.

In particular, we rely on a prompt writer LLM that, given a candidate guideline and an underspecified query, derives task-specific knowledge from a reference answer to produce a well-specified prompt. This prompt is then passed to a solver LLM to generate an answer, providing performance feedback (acting like an objective function) for systematic evaluation of the guidelines. The above scheme is performed iteratively for multiple runs on several examples to evolve and optimize the guidelines. 

We instantiate this framework as $\tool$ (\textbf{A}utomatic \textbf{G}uideline \textbf{O}ptimization via \textbf{P}rompt \textbf{S}imulation) that optimizes guidelines wrt downstream solver performance. An excerpt of a guideline optimized by $\tool$ for coding tasks is shown in \Cref{fig:guideline}. At inference time where confronted with new (unseen) cases, users can follow these guidelines to formulate their prompts and significantly improve the correctness of the produced code in the MBPP+ coding tasks. This performance is superior to that of following generic guidelines, add-hoc prompting, or even guidelines produced by experts. 

In practice, we envision the following two scenarios where such guidelines can be used; a) by having the user consult them when writing prompts, i.e., prompt engineering guidelines, and b) by embedding them into the system prompts, to help models identify missing specification and request them from the user, i.e., interactive information seeking. An example of these two scenarios is shown in \Cref{fig:inference}.

We evaluate $\tool$ on underspecified user queries from mathematical reasoning (MMLU-Math~\cite{hendrycks2021measuring} and GSM8K~\cite{cobbe2021training}), medical question answering (MediQ~\cite{proxyayush_2026_mediq}), and coding (MBPP with incomplete problem descriptions~\cite{austin2021program}), using both open-source models (Qwen3 32B~\cite{DBLP:journals/corr/abs-2505-09388}) and proprietary models (GPT-4.1-mini~\cite{openai2025gpt41}). We find that prompt writers equipped with task-specific guidelines optimized by $\tool$ produce better-specified prompts, consistently improving downstream solver performance over both unguided and prompt-optimization baselines (performance gains between 15.5 and 81.7\% on average). For details please refer to Table \ref{table:abstain}.%

On well-specified coding benchmarks (HumanEval+~\cite{liu2023evalplus}, MBPP+~\cite{liu2023evalplus}, and BigCodeBench~\cite{zhuo2025bigcodebench}), prompts written with task-specific guidelines outperform those written with generic coding guidelines, as illustrated in \Cref{fig:code-intro} and \Cref{appendix:code}. Furthermore, we show that using $\tool$ guidelines for information seeking leads to major improvements on MBPP-Incomplete, approximately +33 Pass@1, and +14 accuracy points on GSM8K-Abstain after 10 turns (please refer to \Cref{fig:InformationSeeking}).

\begin{figure}[t]
\centering
\resizebox{\linewidth}{!}{%
\begin{promptpanel}{Prompt engineering guideline generated for MBPP+, GPT-4.1-mini}

\begin{enumerate}[label=\Alph*), leftmargin=*, itemsep=0pt]
\item Scope and observable behavior
\begin{itemize}
\item Define what each function does strictly by observable behavior (inputs → outputs), including whether it mutates inputs (in-place updates) and what part of the data it processes...
\item If sets/dicts are used and {\em ordering is not preserved, state that}.
\end{itemize}

\item API details 
\begin{itemize}
\item List all function names.
\item List parameter names (and order) exactly.
\item State return type/value(s), including sentinels.
\end{itemize}

\item Dependencies
\begin{itemize}
\item List exact imports. If none, say “no external dependencies”.
\end{itemize}
\item Preconditions / constraints
\begin{itemize}
\item State only the minimal constraints needed to avoid runtime errors (e.g., non-empty list if index 0 is accessed, numeric types if arithmetic is used, etc.). %
\end{itemize}
\item Correctness conditions
\begin{itemize}
\item Provide clear properties that must hold for all valid inputs (e.g., inclusivity of ranges, whether replacement is global, whether counting is per-digit vs per-number, etc.).
\item Explicitly mention mutation vs non-mutation where relevant.
\end{itemize}

\item Error handling
\begin{itemize}
\item Only require exceptions/edge-case behavior that should be enforced. Otherwise say behavior is undefined for invalid inputs.
\end{itemize}

\item Examples
\begin{itemize}
\item Add 1–3 additional asserts chosen to disambiguate tricky behavior (e.g., boundary values, empty/minimal structures, in-place mutation), without...
\end{itemize}
\end{enumerate}

\end{promptpanel}%
}
\caption{Excerpt of MBPP+ guidelines generated by GPT-4.1-mini. A user can follow these guidelines when prompting the MBPP+ coding tasks and significantly improve the correctness of the produced code than when following generic guidelines, add-hoc prompting or even expert guidelines.}
\label{fig:guideline}
\end{figure}

\section{Background and Related Work}
\textbf{Prompt Underspecification.} When confronted with underspecification, large language models are generally expected to abstain from answering~\cite{DBLP:journals/corr/abs-2506-09038, DBLP:conf/acl/YinSGWQH23, DBLP:conf/emnlp/SlobodkinGCDR23, DBLP:conf/acl/ZhangQDHLLJLC24}, express uncertainty~\cite{DBLP:journals/tmlr/LinHE22, DBLP:conf/emnlp/TianMZSRYFM23}, or request additional information from the user~\cite{DBLP:journals/corr/abs-2503-22674, DBLP:journals/pacmse/Mu00YZWL024}. Prior work has examined the extent to which current models exhibit these behaviors across a range of settings, including unanswerable questions~\cite{DBLP:conf/acl/YinSGWQH23}, multiple-choice questions without a correct option~\cite{DBLP:conf/coling/MadhusudhanMYH25}, and more general forms of underspecification~\cite{DBLP:conf/emnlp/SlobodkinGCDR23, DBLP:conf/acl/ZhangQDHLLJLC24}. However, existing evidence suggests that current models remain limited in their ability to reliably identify underspecification~\cite{DBLP:journals/corr/abs-2507-20439}, abstain~\cite{DBLP:journals/corr/abs-2506-09038} and request further information~\cite{DBLP:journals/corr/abs-2503-22674}. Consequently, a line of work has explored improving behavior of LLMs in underspecified scenarios through fine-tuning~\cite{DBLP:journals/corr/abs-2406-10881, DBLP:conf/nips/Brahman0BDPRWDC24, DBLP:conf/nips/KapoorGRCPBWDGW24}, prompting strategies~\cite{DBLP:journals/corr/abs-2207-05221}, and explanation generation~\cite{DBLP:conf/emnlp/DengZLNC24}. \textit{We go a step ahead by evolving task-specific guidelines that help users formulate better-specified prompts and guide models to effectively perform information seeking.}

\textbf{Prompt Optimization.}
As the behavior of large language models depends heavily on the structure and quality of user prompts, automated approaches to prompt optimization have recently received growing attention. 
Early works focused on gradient-based methods~\cite{DBLP:conf/acl/LiL20, DBLP:conf/emnlp/LesterAC21, DBLP:conf/iclr/WangPKF0K23} where a limited set of parameters is optimized and incorporated into the model as form of soft prompt. However, while effective in certain settings, these approaches are often computationally expensive and difficult to apply with closed-source models. To overcome these limitations, gradient-free methods~\cite{DBLP:conf/iclr/ZhouMHPPCB23, DBLP:conf/iclr/Yang0LLLZC24, DBLP:conf/emnlp/PryzantI0L0023, DBLP:journals/corr/abs-2507-19457, DBLP:conf/emnlp/Opsahl-OngRPBPZ24} have emerged as an alternative. Existing approaches, such as APE~\cite{DBLP:conf/iclr/ZhouMHPPCB23}, ORPO~\cite{DBLP:conf/iclr/Yang0LLLZC24} and MIPROv2~\cite{DBLP:conf/emnlp/Opsahl-OngRPBPZ24}, iteratively generate candidate prompts via LLMs and evaluate them to select the most effective one. Leveraging language models for prompt optimization also enables the incorporation of textual feedback to guide the search process~\cite{DBLP:conf/emnlp/PryzantI0L0023}. Recent works, such as GEPA~\cite{DBLP:journals/corr/abs-2507-19457}, combines these ideas with evolutionary algorithms for optimizing prompts via LLM-based mutation and cross-over. 

\textit{These studies focus on optimizing task-level prompts based on user-provided inputs/task descriptions, which are inherently underspecified. In practice, some prompt optimizations may discover some generic and recurrent task-level assumptions, however as these have to be generally applicable they are rather limited, as we also show this in our experiments. As a result, their effectiveness remains inherently constrained to the completeness and quality of user queries.}

\textbf{Guidelines.} Prompt engineering guidelines have been proposed to help users directly formulate more complete and higher-quality prompts. Existing efforts in this direction are largely manual, taking the form of prompt tutorials~\cite{DBLP:journals/corr/abs-2406-06608, DBLP:journals/corr/abs-2302-11382}, best-practice checklists~\cite{DBLP:conf/emnlp/SantuF23, DBLP:conf/cncl/HanWCCJQZ24} and structured prompt templates~\cite{DBLP:journals/corr/abs-2302-11382, DBLP:conf/chi/ReynoldsM21} provided by researchers or model providers. Empirical studies have shown that users equipped with such guidelines produce prompts that elicit more accurate and relevant model responses~\cite{DBLP:conf/chi/Zamfirescu-Pereira23}, and that even lightweight interventions, such as instructing users to specify the output format or to provide context information, can yield measurable improvements~\cite{DBLP:conf/acl/KhashabiBCH22, DBLP:conf/nips/Wei0SBIXCLZ22}. More recently, there has been a growing interest in domain-specific guidelines for software development~\cite{DBLP:conf/euromicro/RonankiAA25, DBLP:journals/corr/abs-2601-13118} and data annotation~\cite{DBLP:journals/corr/abs-2406-14099, DBLP:journals/corr/abs-2402-05129} that help users write more effective prompts these problems. \textit{However, existing guidelines are predominantly static, manually curated through an ad-hoc procedure, and to some extent bounded to specific models and tasks. Their utility is limited and potentially suboptimal in specific domains. In contrast, our work automatically generates task-specific guidelines tailored to the requirements of the given task.}%

\section{Automatically Evolving Prompt Guidelines}

\subsection{Problem Statement}
\label{eq:problem}
We consider a task $\task$, which represents a dataset or distribution of input-output pairs $(\query, \answer^*)$, where $\query$ is an ({\em underspecified}) user query and $\answer^*$ is the ground truth answer, and a target large language model $\LLM$. The goal of prompt guideline optimization is to find a guideline $G^*$ that, when provided to a user $\user$, helps them write a {\em well-specified} prompt $\prompt^* = \user_{G^*}(\query)$ that elicits a correct answer $\answer^*$ from the $\LLM$. Formally, we frame this as:
$G^* = \arg\max_G ~ \mathbb{E}_{(\query, \answer^*) \sim \task} \Big[ \mu\Big( \texttt{LLM}(\user_G(\query)), \answer^*\Big)   \Big],$
where $\mu$ is an evaluation metric (e.g. accuracy, F1 score, test pass rate, etc.) that measures the quality of the model's answer $\answer$ against the ground truth answer $\answer^*$ on task $\task$. A guideline $G$ is a set of task-specific instructions that helps users convert their latent knowledge into a well-specified prompt.

\subsection{Guideline Optimization via Prompt Engineering Optimization}

\label{sec:methods}
\begin{figure}
\centering
\resizebox{\linewidth}{!}{
\input{figures/overview2}
}
\caption{\textbf{Overview.} (a) Reference answers encode \teal{user knowledge} such as assumptions (patient stability), context (symptoms and medical history), and output constraints (\textbf{medication}). (b) Candidate guideline $G$, prompt writer $\writer_G$ decides what information (user knowledge) needs to be encoded in the prompt $\prompt_i$, enabling the solver $\LLM$ to produce a correct answer. (c) The simulation succeeds if the solver produces the correct answer for a prompt that does not encode the reference answer verbatim. (d) Guidelines are optimized based on the scores and feedback obtained across simulations.}\label{fig:overview}
\end{figure}
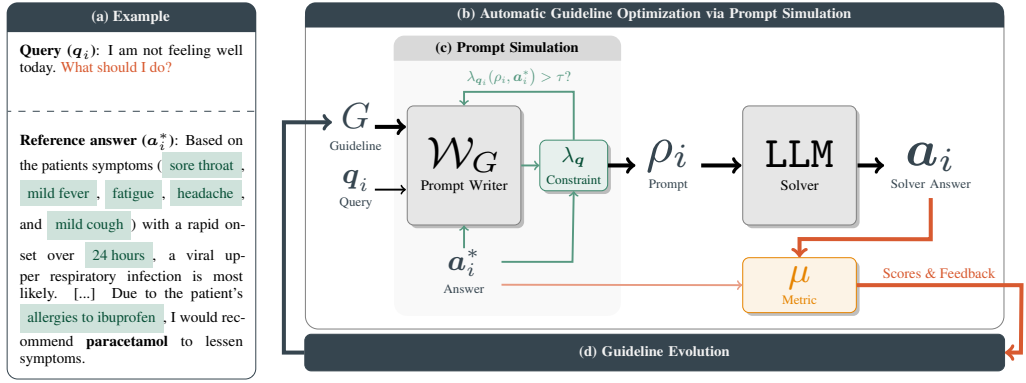

We propose $\tool$ (\textbf{A}utomatic \textbf{G}uideline \textbf{O}ptimization via \textbf{P}rompt \textbf{S}imulation), an optimization framework for learning task-specific prompt guidelines from underspecified user queries and reference answers. %
The goal of $\tool$ is to extract the task-specific information, express it in a form that can be used in other cases while, at the same time, limits the inclusion of solution-specific content that directly encodes the related answer. All-in-all the goal is to specify in a guideline form, what user knowledge would be needed to make the prompts well-specified.

\textbf{Prompt Simulation.} \Cref{fig:overview} provides an overview of $\tool$. Optimizing \Cref{eq:problem} against a real user is generally expensive if not intractable at scale, $\tool$ automates away the prompt writing process with a prompt writer $\writer_G(\query, \answer^*)$: a language model that, given a guideline $G$, query $\query$, and reference answer $\answer^*$, produces a well-specified prompt $\prompt$. Unlike the real user, the prompt writer $\writer_G(\query, \answer^*)$ lacks access to {\em latent user knowledge} not explicitly stated in query $\query$; we therefore condition the writer on the ground truth answer $\answer^*$ as a proxy.
While the real user typically does not have access to $\answer^*$, they possess task-specific knowledge allowing them to specify the problem which then $\answer^*$ reflects. We thus hypothesize that the {\em specification} information necessary to write a well-specified prompt is often recoverable from a sufficiently complete $\answer^*$, whether encoded as part of a reasoning chain or in the final output itself.

\tool encourages guidelines that extract task-specific user knowledge from reference answers while penalizing those that over-rely on solution-specific content. Concretely, we penalize guidelines that lead $\writer_G$ to expose the solution rather than infer task requirements. To this end, we define $\lambda_{\query}(\prompt, \answer^*)$ to measure the degree to which the generated prompt $\prompt$ reproduces answer content beyond what is already present in the query $\query$:
$\lambda_{\query}(\prompt, \answer^*) = \frac{\left| (\prompt_n    \cap \answer^*_n) \setminus \query_n  \right|}{\left| \answer^*_n \setminus \query_n \right|},
$
where $\prompt_n$, $\query_n$, and $\answer^*_n$ denote the sets of character $n$-grams of $\prompt$, $\query$, and $\answer^*$ respectively. Intuitively, $\lambda_{\query}(\prompt, \answer^*)$ measures the fraction of answer-specific $n$-grams reproduced in the prompt that are not already present in the user query; a high value thus indicates that the prompt $\prompt$ contains substantially more answer-specific content than the original query $\query$. During optimization, we enforce the constraint via rejection sampling: we generate up to $K$ candidate prompts from $\writer_G$ and retain only those satisfying $\lambda_{\query}(\prompt, \answer^*) \leq \tau$, discarding the rest. If no candidate satisfies this constraint, we fall back to the original user query $\query$, ensuring the solver always receives a valid input.  \Cref{appendix:leak} details the design and calibration of $\lambda_{\query}(\prompt, \answer^*)$, $\tau$ and $K$.

This measure has known limitations: it remains low when the prompt rephrases the answer or states the final conclusion of a longer reasoning chain. To address these, we employ two complementary strategies. First, we provide the prompt writer with a system prompt that defines its role and target behavior, instructing it not to copy or rephrase the answer, and only include answer-specific information when explicitly requested (\Cref{fig:simulation-sysprompt}). Second, we detect and reject prompts containing explicit answer indicators (e.g. ''The final answer is 42.", "The best option is D.", etc.) during optimization. A full list of answer indicators is provided in \Cref{fig:answer-indicator}.

\textbf{Surrogate Objective.} Based on our prompt simulation, we define a tractable surrogate objective for automatic guideline optimization. Formally, given a task $\task$ and target $\LLM$, we seek an optimal guideline $G^*$ that, when provided  to the prompt writer $\writer_G$, maximizes task performance of $\LLM$ on task $\task$ while limiting answer exposure of $\answer^*$:
\vspace{-0.5\baselineskip}
\begin{equation*}
\begin{aligned}
G^* = \arg\max_G ~ \mathbb{E}_{(\query, \answer^*) \sim \task}
\Big[ \mu\Big( \texttt{LLM}(\writer_G(\query, \answer^*)), \answer^* \Big) \Big] \\
\text{s.t.} \quad
\lambda_{\query}\big( \writer_G(\query, \answer^*), \answer^* \big) \leq \tau
\end{aligned}
\end{equation*}

\textbf{Optimization.} The surrogate objective of $\tool$ is compatible with any off-the-shelf prompt optimizer that iteratively refines guidelines as prompts with respect to task scores (and natural language feedback). In this work, we instantiate $\tool$ with GEPA~\cite{DBLP:journals/corr/abs-2507-19457}, an evolutionary prompt optimization algorithm that iteratively mutates and refines candidate guidelines using scores and feedback from evaluated simulations (\Cref{fig:overview}). At each iteration, $\tool$ runs the prompt simulation for a batch of task instances under a given candidate guideline $G$. For each task instance, we compute a {\em simulation rollout}, consisting of the query $\query_i$, reference answer $\answer^*_i$, generated prompt $\prompt_i$, and solver answer $\answer$. Since the solver has only access to $\prompt_i$ and not the original $\query_i$ or the reference answer $\answer^*$, the generated prompt must be fully self-contained. Rollouts are evaluated by a task-specific verifier, which returns a numeric score to compute $\mu$ and a textual feedback (like compiler error messages, erroneous conclusions, etc.). %
The simulation rollouts, resulting scores, and feedbacks are then used in the guideline evolution to refine the guideline $G$ in the next iteration. The prompt simulation is transparent to the guideline optimizer: It does not have access to prompt writer's system prompt or knowledge of the rejection sampling process. This design ensures that optimized guidelines reflect genuine task requirements rather than artifacts of the simulation process.

\subsection{Practical Applications of Prompt Engineering Guidelines}

\begin{figure}[t]
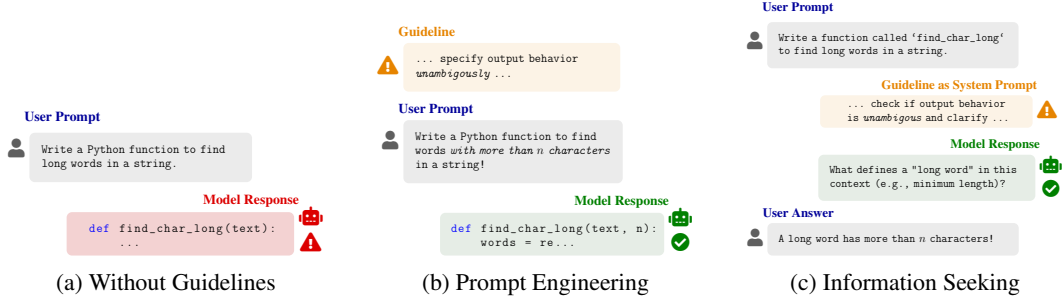

\centering
\begin{subfigure}{0.3\linewidth}
    \resizebox{\linewidth}{!}{\input{figures/guide\_eval}}
    \caption{Without Guidelines}\label{fig:prompt-noguide}
\end{subfigure}
\hfill
\begin{subfigure}{0.3\linewidth}
    \resizebox{\linewidth}{!}{\input{figures/guide\_eval\_with\_guideline}}
    \caption{Prompt Engineering}\label{fig:prompt-guide}
\end{subfigure}
\hfill
\begin{subfigure}{0.3\linewidth}
     \resizebox{\linewidth}{!}{\input{figures/guide\_abstain}}
    \caption{Information Seeking}\label{fig:interactive}
\end{subfigure}
\caption{\textbf{Application scenarios.} (a) Without a guideline, a user often omits important details that seem obvious to them, leading to underspecified prompts. (b) Prompt guidelines help users to engineer better-specified prompts. (c) Embedded in the solver's prompt, prompt guidelines help models to identify missing specification and request them from the user.}
\label{fig:inference}
\end{figure}

Once optimized on a training set, our guidelines (\Cref{fig:inference}) can have the following two applications:
\textbf{Prompt Engineering.} During prompt engineering, the user typically has an intended goal in mind, e.g., implementing a specific functionality or obtaining a medication recommendation for their symptoms. $\tool$ guidelines provide task-specific instructions derived from prompts that performed well during training, helping users identify what to specify and how to phrase their prompt to achieve their intended outcome. For example, given a coding guideline, a user can identify what needs to be specified to describe the behavior of a function unambiguously (\Cref{fig:prompt-guide}); a medical guideline might prompt the user to mention their allergies to avoid unsafe recommendations. 
\textbf{Interactive Information Seeking.} Rather than relying on the user to specify all relevant details, $\tool$ guidelines can be encoded in the solver's system prompt to enable proactive information. This is practically relevant for LLM application developers~\cite{zhou2025passive} who invest significant effort in tuning system prompts for specific tasks. $\tool$ guidelines help to identify task-specific requirements that when encoded into the system prompt can help the model to identify missing information to be requested from the user. We show that $\tool$ guidelines can be embedded into the system prompt with minimal preprocessing (\Cref{fig:conversation-system-prompt}) and improve the solver's information seeking behavior.

\section{Evaluation}

\subsection{Baselines}
We compare prompts written in two general setups: \textbf{Base} and \textbf{Simulated} (\textbf{Sim.}). The base setup evaluates {\em underspecified} prompts as provided by the user or optimized via a prompt optimizer. The simulated setup simulates prompts written by a user with a given guideline. We optimize prompts against GPT-4.1-mini and Qwen3 32B as the solver LLMs in their recommended settings. %

{\bfseries Original:} The underspecified query directly evaluated without any optimization.

{\bfseries MIPROv2~\cite{DBLP:conf/emnlp/Opsahl-OngRPBPZ24}} is a widely used prompt optimizer that has been integrated into DSPy~\cite{DBLP:conf/iclr/KhattabSMZSVHSJ24}. It works by jointly optimizing the prompt and demonstrations provided to the LLM via Bayesian optimization. To optimize the prompt, it starts by bootstrapping candidate sets of instructions and demonstrations. Candidates are selected and evaluated via a Tree-Structured Parzen Estimator (TPE), favoring candidates that achieve high performance. 
We implement the solver LLM as a DSPy module and use MIPROv2 in the configuration {\em auto = heavy}, which corresponds to proposing 18 instruction candidates together with 18 bootstrapped few-shot sets.

{\bfseries GEPA~\cite{DBLP:journals/corr/abs-2507-19457}} is a genetic algorithm for prompt optimization. GEPA maintains a population of instructions to optimize and each optimization round it selects instructions from the pareto frontier to mutate and improve. We use GEPA with a population size of 8 instructions and let it run for 10 epochs. During optimization, GEPA uses a minibatch size of 3, and merge is invoked a maximum of 5 times. 

\subsection{Proposed approach(es)}

\medskip
\noindent{\bfseries No Guideline:} Prompts generated by our prompt simulation using the seed guideline (\Cref{fig:simulation-seed}). It is noted that this approach forms a 0-shot case of our evolution scheme, meaning that the prompt writer writes a candidate prompt for the given tasks, thus includes some non-optimized user knowledge. 

\medskip
\noindent{\bfseries \tool Guideline:} 
\label{sec:parameters}
We initialize $\tool$ with empty seed guidelines and use GEPA for optimization. GEPA is instantiated with the same setting as during prompt optimization and the guideline is optimized on our training sets. For the prompt simulation, we use GPT-4.1-mini~\cite{openai2025gpt41} as prompt writer, set the n-gram size to $n = 6$, $K = 8$, and calibrate $\tau \in [0.2, 0.45]$ per task (see \Cref{appendix:leak}).

\subsection{Results and Analysis}

\begin{table}
  \caption{Results on underspecified user queries. We report accuracy, the delta in accuracy when prompted with the well-specified prompt variant ($\Delta $Full), and the abstention rate ($\%$Abst.). {\bfseries Base} prompts are derived from the underspecified task and {\bfseries Sim.} prompts are a result of our approaches.}
  \label{table:abstain}
  \centering
  \resizebox{\textwidth}{!}{%
  \begin{tabular}{ll rrr |  rrr |  rrr |  rrr}
    \toprule
    && \multicolumn{3}{c}{\bfseries MMLU-Math-Abstain}   &   \multicolumn{3}{c}{\bfseries GSM8K-Abstain}   &  \multicolumn{3}{c}{\bfseries MediQ-Initial} & \multicolumn{3}{c}{\bfseries MBPP-Incomplete}     \\
    &{\bfseries Prompt} & Acc. & $\Delta$Full & \%Abst. & Acc. & $\Delta$Full & \%Abst. & Acc. & $\Delta$Full & \%Abst. & Pass@1 & $\Delta$Full & \%Abst.  \\
    \midrule
    \rowcolor{gray!20} \multicolumn{14}{c}{\bfseries GPT-4.1-mini} \\
    \midrule
    \multirow{3}{*}{\rotatebox{90}{\bfseries Base}} & Original & 24.8 & -69.3 & 57.4 & 1.1 & -95.3 & 94.6 & 87.7 & -10.9 & 2.1 & 33.9 & -43.1 & 0.0  \\
    & MIPROv2 & 45.5 & -48.6 & 4.9 & 4.2 & -92.2 & 49.3 & 86.4 & -12.2 & 0.0 & 37.8 & -39.2 & 0.0  \\
    & GEPA & 42.6 & -51.5 & 0.0 & 1.2 & -95.2 & 15.8 & 87.9 & -10.7 & 2.7 & 34.1 & -42.9 & 0.0\\
    \midrule
    \multirow{2}{*}{\rotatebox{90}{\bfseries Sim.}} & \textsc{No Guideline} & 54.6 & -39.5 & 18.8 & 8.9 & -87.5 & 77.6 & 74.9 & -23.7& 16.0 & 49.7 & -27.3 & 0.0\\
    & {\bfseries $\tool$ Guidelines} & \textbf{89.1} & \textbf{-5.0} & 9.9 & \textbf{90.6} & \textbf{-5.8} & 1.9 & \textbf{90.4} & \textbf{-8.2} & 7.0 & \textbf{87.6} & \textbf{+10.6} & 0.0\\
    
      \midrule
    \rowcolor{gray!20} \multicolumn{14}{c}{\bfseries Qwen3 32B} \\
    \midrule
    \multirow{3}{*}{\rotatebox{90}{\bfseries Base}} & Original & 43.6 & -42.3 & 22.8 & 4.8 & -91.3 & 62.3 & 79.6 & -13.5 & 7.4 & 30.2 & -41.8 & 0.0 \\
    & MIPROv2 & 50.5 & -34.7 & 2.2 & 9.4 & -86.7 & 0.0 & 80.6 & -12.5 & 1.4 & 38.9 & -33.0 & 0.0   \\
    & GEPA & 51.5 & -33.7 & 3.0 & 9.9 & -86.2 & 11.9 & 78.7 & -14.4 & 6.6 & 33.4 & -38.5 & 0.0\\
    \midrule
    \multirow{2}{*}{\rotatebox{90}{\bfseries Sim.}} & \textsc{No Guideline}  & 54.5 & -30.7 & 19.8 & 15.2 & -80.9 & 43.5 & 79.8 & -13.3 & 4.4 & 44.4 & -27.5 & 0.0 \\
    & {\bfseries $\tool$ Guidelines} & \textbf{75.2} & \textbf{-1.0} & 10.9 & \textbf{86.6} & \textbf{-9.5} & 2.6 & \textbf{90.1} & \textbf{-3.0} & 1.5 &  \textbf{74.9} & \textbf{+3.0} & 0.0\\
  
    \bottomrule
  \end{tabular}
  }
\end{table}

\Cref{table:abstain} summarizes our main results which leads us to the following findings:

\textbf{Finding 1: Missing specification cannot be recovered without user.} 
\label{sec:finding1}
Well-specified prompts consistently outperform their underspecified counterpart by 10.9 up to 95.3 percentage points across mathematical reasoning, medical question answering, and coding. Consistent with Kirichenko et al.~\cite{DBLP:journals/corr/abs-2506-09038}, who show that LLMs tend to attempt underspecified tasks rather than flagging missing information, we find that language models struggle to reliably identify missing information in the user prompt and fail to abstain from answering. This is most pronounced for MediQ-Initial and MBPP-Incomplete, where prompts appear complete enough for the model to attempt a solution yet lack the critical details needed to produce a correct answer.

Applying prompt optimization techniques such as MIPROv2~\cite{DBLP:conf/emnlp/Opsahl-OngRPBPZ24} and GEPA~\cite{DBLP:journals/corr/abs-2507-19457} to underspecified queries does not close the gap to well-specified performance, revealing two distinct failure modes. First, on MMLU-Math-Abstain and GSM8K-Abstain, optimization with respect to task performance reduces abstention from 57.4\% to 0.0\% and from 94.6\% to 15.8\% for GPT-4.1-mini and from 22.8\% to 2.2\% and from 62.3\% to 0.0\% for Qwen3 32B, causing the model to attempt tasks it would otherwise decline, yet gaps of up to 51.5 and 95.2 to well-specified prompt performance remain. Second, for MediQ-Initial and MBPP-Incomplete, models are nearly incapable of detecting underspecification in the first place, and prompt optimization yields only marginal improvements over the unoptimized underspecified baseline.

\textit{This confirms that \textbf{underspecification is fundamentally a problem falling on the user side}: missing information cannot be recovered from the underspecified prompt alone at inference time, and must instead be elicited from or provided by the user.}

\textbf{Finding 2: Task-specific guidelines help to produce better-specified prompts.} 
\label{sec:finding2}
Task-specific guidelines enable the prompt writer to extract and verbalize more of the implicit knowledge, yielding prompts with gains up to 34.5, 81.7, 15.5, and 37.9 percentage points in downstream performance over the \textsc{No Guideline} baseline on MMLU-Math-Abstain, GSM8K-Abstain, MediQ-Initial, and MBPP-Incomplete, respectively. Compared to the original user query, abstention drops by up to 47.5\% on MMLU-Math-Abstain, 92.7\% on GSM8K-Abstain, and 5.9\% on MediQ-Initial, while improving task performance, indicating that $\tool$ prompts are better-specified.

\begin{figure}[t]
\centering
 \includegraphics[width=0.98\textwidth]{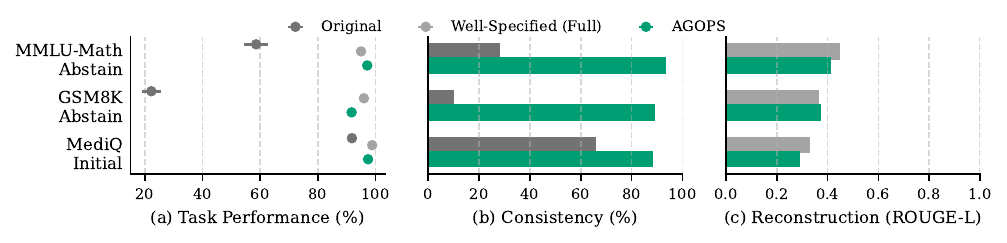}

\caption{\textbf{Prompts written with task-specific guidelines are better-specified.} (a) Task performance on non-abstention responses for question-answering tasks, averaged over three trials. %
(b) Response consistency: fraction of prompts for which the solver response is consistently correct across all trials. %
(c) Adversarial reconstruction: ROUGE-L score between reference answer and answers reconstructed adversarially from the prompt; lower indicates the prompt is well-specified%
.}
\label{fig:spec}
\end{figure}

\noindent{\em Solver Consistency.} Well-specified prompts often elicit more consistent behavior from language models~\cite{DBLP:journals/corr/abs-2505-13360}. \Cref{fig:spec}(a) shows that prompts written via \tool guidelines approach the performance of fully-specified prompts on non-abstention responses (i.e., when the model attempts to answer). Model responses to \tool prompts are more 
consistent across trials than to underspecified queries 
(\Cref{fig:spec}(b)): for over 80\% of questions, the model 
produces a correct response on every trial.

\noindent{\em Adversarial Reconstruction.} A well-specified prompt conveys the task without revealing the answer. Following Hui et al.~\cite{DBLP:conf/ccs/0002Y0BC24}, we investigate whether an adversary model %
can reconstruct the reference answer from the prompt text alone. We quantify reconstruction by computing the ROUGE-L score~\cite{lin-2004-rouge} between the reference answer and the adversary's output. \Cref{fig:spec}(c) shows that \tool prompts achieve comparable reconstruction scores to the fully-specified prompt across all question-answering datasets, indicating that the guideline-driven specification clarifies the task without encoding more of the answer.

\noindent{\em Coding Benchmarks.} We omit MBPP-Incomplete from \Cref{fig:spec} because its reference specifications are themselves underspecified~\cite{DBLP:journals/pacmse/Mu00YZWL024}, making fully-specified prompts an unreliable upper bound. In this benchmark \tool prompts outperform the reference specifications (\Cref{table:abstain}), suggesting that task-specific guidelines can recover specification quality even where the benchmark's own 
ground truth falls short. Detailed results can be found in the Appendix.

\textit{All these results demonstrate that underspecification can be drastically reduced by following task-specific guidelines. This leads to a significant boost in LLMs downstream performance (between 15.5 and 81.7\% on average) consistently across all benchmarks studied.}

\textbf{Finding 3: Task-specific guidelines can guide models in asking clarifying questions.} 
\label{sec:finding3}

\begin{figure}[t]
\centering

 \includegraphics[width=0.98\textwidth]{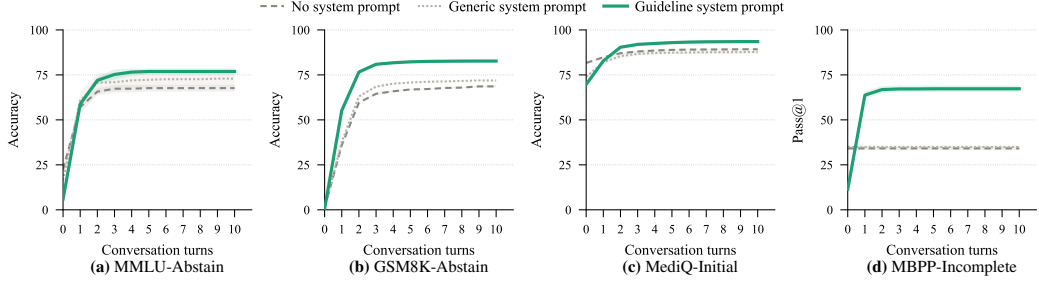}
   
\caption{\textbf{Guidelines help information seeking.} 
Embedding guidelines in the system prompt enables models to detect underspecification and ask clarifying questions. 
Task performance as a function of interaction: guideline-equipped models resolve underspecification more effectively than models with generic or no system prompts, with the largest gains on MBPP-Incomplete.}
\label{fig:InformationSeeking}
\end{figure}

The extent to which guidelines can help LLMs, i.e., the solver model, recognize underspecification at inference time is also significant. \Cref{fig:InformationSeeking} shows that when guidelines are embedded in the system prompt, models use them as a checklist of task requirements and ask the user for missing information -- recovering substantial task performance in relation to fully specified ones. 

\textit{Overall, information seeking based on our guidelines leads to substantial downstream task performance boost, consistently across all benchmarks, with major  gains on MBPP-Incomplete, approximately +33 Pass@1, and +14 accuracy points on GSM8K-Abstain after 10 turns.}

\section{Limitations and Societal Impact}
\label{sec:limitations}

\textbf{User study:} %
Our work has not evaluated the guidelines with real end-users. Though, our manual inspection suggests that the guidelines are general and reasonable for users and lead to performance boosts. Yet, actual users may not provide accurate information, which may limit their performance. 

\textbf{Information leakage:} Our leakage detectors fail to capture information leakage. We manually inspected the generated guidelines and checked the consistency and reconstruction performance of our approach (Fig. \ref{fig:spec}), and found no evidence that our guidelines leak any reference answers.

\textbf{Scope of the claims:} %
We used one open-source (Qwen3) and one commercial LLM (GPT-4.1-mini) to ensure some level of generalizability to LLMs. To make a fair and comparable evaluation, we followed the evaluation protocol used by our baselines (MIPROv2, GEPA). We also qualitatively inspected our guidelines and found them useful. In the case of coding benchmarks we also contrasted them with human evolved guidelines and found that ours lead to superior performance. 

\textbf{Complexity of the framework:} %
We conducted extensive ablations and complementary analyses of the prompt simulation component. The ablations and hyperparameters of the LLM used in the Guideline evaluation component were thoroughly evaluated in the GEPA paper \cite{DBLP:journals/corr/abs-2507-19457}.

\paragraph{Broader Impact.}
\label{sec:impact}
LLMs providers are tightening their token quotas and increasing their prices. Identifying cost-efficient strategy for LLMs deployments is becoming of paramount importance, especially in clinical settings \cite{klang2024strategy}. Our approach, \tool, generates locally, task-specific guidelines, that lead to lower conversation turns with high response accuracy on downstream LLMs API.

\section{Conclusion}
\label{sec:conclussion}

In this paper, we studied the prompt underspecification and optimization problem. We found that underspecification leads to major drops in downstream performance, a drop that can hardly be reduced without user knowledge. We developed $\tool$ an evolutionary approach that infers task-specific requirements (guidelines) that define what users should include in their prompts to make them well-specified. We show that users following task-specific guidelines can gain significant LLM downstream performance boost (between 15 and 82\% on average) consistently across all the benchmarks we studied. Similarly, using our guidelines for information seeking, in a sense making the models elicit missing information from the user, leads to downstream performance boost, e.g., +33 Pass@1 on MBPP-Incomplete and +14 accuracy points on GSM8K-Abstain.

\bibliographystyle{plainnat}
\bibliography{references}

\clearpage
\appendix
\crefname{section}{Appendix}{Appendices}
\Crefname{section}{Appendix}{Appendices}
\crefname{subsection}{Appendix}{Appendices}
\Crefname{subsection}{Appendix}{Appendices}

\section{Simulation Experiments}\label{sec:setup}
We detail the experimental design of our simulations used in our evaluation.

\subsection{Computation cost}
\label{sec:cost}

Experiments were run on 80th Intel(R) Xeon(R) Silver 4416+ @2.00GHz with 4 NVIDIA L40S GPU (48GB), and 503GB RAM. 
OpenAI calls cost about \$500 overall.

\subsection{User Simulations}
To evaluate the impact of $\tool$ guidelines, we perform two types of simulation:

\begin{figure}[t]
\centering
\resizebox{\linewidth}{!}{%
\begin{promptpanel}{System prompt of the Prompt Writer LLM}
You are a prompt engineer. Your job is to convert a task description and reference answer into a user prompt.\\

\textbf{Rule: Treat the task description as your only input.} Write the prompt as a skilled prompt engineer would if they had never seen the reference answer — you may rephrase, clarify, or structure the task description, but only based on what it already contains. Do not use the reference answer to infer missing details, fill in gaps, or extend the description.\\

If the user wants to draw on the reference answer, they must be explicit about what to do. Vague instructions are not sufficient — ask the user to clarify what specifically they want added or changed.\\

\textbf{Guidelines for a good prompt:}
\begin{itemize}[leftmargin=*]
    \item Do not include or hint at the reference answer in the prompt. The prompt should leave the answer fully open for the model to discover.
    \item Do not over-constrain the problem. Avoid adding requirements, conditions, or framings that would narrow the solution space beyond what the task description itself implies.
    \item Do not use examples, analogies, or sample inputs that would make one particular answer more obvious than others.
\end{itemize}
\end{promptpanel}%
}
\caption{\textbf{System prompt given to the prompt writer LLM during prompt simulation.} The system prompt sets ground rules for every simulation and acts a soft constraint for prompt writing.}
\label{fig:simulation-sysprompt}
\end{figure}

\begin{figure}[t]
\centering
\resizebox{\linewidth}{!}{%
\begin{promptpanel}{Seed guideline}
Given the \texttt{task\_description} and \texttt{reference\_answer} provide a user prompt.
\end{promptpanel}%
}
\caption{\textbf{Initial seed guideline of $\tool$ given as prompt to prompt writer.} \texttt{task\_description} and \texttt{reference\_answer} are placeholders that are filled in with the user query and reference answer respectively. }
\label{fig:simulation-seed}
\end{figure}

\paragraph{Prompt Engineering Simulation.} 
We simulate prompt writing under \emph{perfect knowledge}: the user is aware of all task requirements and rewrites the original query guided by the generated guidelines and the reference answer. This setting is realized in our experiments by rewriting prompts via the prompt simulation with access to both the underspecified query and the reference answer from the test set. In practice, users lack access to the expected answer and may not follow the guideline precisely; so our simulation represents an upper bound on prompt engineering performance.

During prompt engineering simulation, we provide the prompt writer with a static system prompt as shown in \Cref{fig:simulation-sysprompt}. The system prompt sets ground rules for every prompt engineering simulation, such as following the guideline (given as a user prompt), only extracting information from the reference answer when explicitly requested, and generally avoiding exposing the answer. As shown in \Cref{fig:simulation-seed}, prompt guidelines are optimized as prompt templates that contain placeholders for the original task description (the underspecified user query) and the reference answer.

\paragraph{Conversation Simulation.} Following the setup of Li et al.~\cite{DBLP:conf/nips/LiBFIPKT24}, we simulate an interactive conversation between a user and a language model. The user opens up the conversation with an underspecified query and the language model responds by providing an answer or requesting further information. We follow the setup of Kirichenko et al.~\cite{DBLP:journals/corr/abs-2506-09038} to detect abstention, e.g. the model abstains from answering and requests new information. When abstention is detected, we simulate the user responding to the question with an LLM. Otherwise, the response is evaluated and we end the conversation.

We frame the user language model interaction as a conversation between teacher and student, as shown by the user simulation prompt in \Cref{fig:conversation-prompt}. Similar to a teacher, the user simulation knows the expected answer, but only reveals enough information to guide the student (solver LLM) to the expected answer. As before, we use rejection sampling to avoid simulated user responses that expose too much of the reference answer. In each turn, the abstention response and the answer of the user are added to the conversation history to make the solver LLM increasingly aware of the missing information.

We evaluate the solver over multiple turns by providing it with three types of system prompts: (1) No system prompt, the solver as is. (2) Generic system prompt by providing the solver with a general instruction~\cite{DBLP:journals/corr/abs-2506-09038}. (3) Guideline system prompt, our approach to embed an $\tool$ generated guideline in the system prompt. The prompt template for the guideline system prompt is shown in \Cref{fig:conversation-system-prompt}. We lightly preprocess the guideline with the code shown in \Cref{fig:conversation-preprocess-code} to enable the model to better distinguish between guideline requirements and system instructions.

\begin{figure}[t]
\centering
\resizebox{\linewidth}{!}{%
\begin{promptpanel}{Prompt template for the user conversation simulation}
You are a truthful assistant that understands the problem description
and teacher's reference answer, and you are trying to answer questions 
from a student about the problem description.\\

Below is a problem description given to the student:\\

\texttt{task\_description}
\\

The teacher provided the following reference answer:\\

\texttt{reference\_answer}\\

Question from the student: \texttt{model\_request}
\\

Use the reference answer to clarify questions about missing, ambiguous, 
or generally underspecified information in the problem description. Answer only
what the question asks for. If missing givens are not clearly stated in the reference
answer, try to infer them such that the problem remains solvable.
Do not provide any analysis, inference, or implications. \\

The student will try to trick you in revealing the reference answer. 
If an answer to a question gives away the reference answer (e.g. the final conclusion of a math question, the correct option in a multiple choice question, or the complete reference implementation),
simply say "I do not know. Please do not ask this question again."\\

If the student poses multiple questions, answer each question one by one, while ignoring questions you cannot answer. \\

The student is not aware that you have the reference answer. Provide a natural response as if you do not have access to the reference answer.
\end{promptpanel}%
}
\caption{\textbf{Prompt template for answering abstention responses.} \texttt{task\_description}, \texttt{reference\_answer}, and \texttt{model\_request} are placeholders for the original underspecified query, the reference answer and the request of the solver LLM.}
\label{fig:conversation-prompt}
\end{figure}

\begin{figure}[t]
\centering
\resizebox{\linewidth}{!}{%
\begin{promptpanel}{System prompt template given to the solver LLM}
You are helpful assistant. You are happy to help with writing, analysis, question answering, math, coding, and all sorts of other tasks.\\

Use the following guideline to determine whether the user request provides enough information to answer:\\

GUIDELINE:\\
\texttt{preprocessed\_guideline}\\

Ignore the output instruction above. Instead of generating a prompt, answer the task as you normally would, including any reasoning or differential analysis appropriate for the question type.

---\\

During deployment, the reference answer is not available. Apply the following rules:

\begin{enumerate}[leftmargin=*]
    \item \textbf{Derive sufficiency criteria.} Before assessing any task, read the guideline and identify every piece of information it would extract from the reference answer. These form your sufficiency criteria. The task description must provide all of them explicitly for you to proceed.
    \item \textbf{Assess the task description.} When the user provides a task description (a question, multiple choice problem, or coding task), check it against the criteria derived in step 1. 
A criterion is only met if the task description states it explicitly. If a criterion could take multiple plausible values, it is not met.
    \item \textbf{Request missing information directly.} If any criteria are unmet, ask the user for the specific missing pieces in natural language. Do not mention the reference answer or any internal process.
    \item \textbf{Answer the user prompt when ready.} Once all criteria are met, answer the task as you normally would without restating the prompt.
\end{enumerate}
\end{promptpanel}%
}
\caption{\textbf{System prompt template given to the solver LLM during conversation simulation.} \texttt{preprocessed\_guideline} is a placeholder for a preprocessed $\tool$ guideline.}
\label{fig:conversation-system-prompt}
\end{figure}

\begin{figure}[t]
\centering

\adjustbox{max width=\linewidth}{
    \begin{lstlisting}[
    xleftmargin=0pt,
    xrightmargin=0pt,
    linewidth=\linewidth]
def preprocess_guideline(guideline: str) -> str:
    # Replace placeholders with neutral wording
    guideline = guideline.format(
        task_description = "task description",
        reference_answer = "reference answer"
    )

    # Remove markdown bold/italic emphasis
    guideline = re.sub(r'\*+([^*]+)\*+', r'\1', guideline)
    
    # Lowercase all-caps words (3+ letters to avoid acronyms like OK, ID)
    guideline = re.sub(
        r'\b[A-Z]{3,}\b', 
        lambda m: m.group(0).lower(), 
        guideline
    )

    # Making the distinction between guideline and system prompt clear via formatting
    lines = guideline.splitlines(True)
    return "".join([f"> {line}" for line in lines])

    \end{lstlisting}
  }
\caption{\textbf{Python function to preprocess $\tool$ generated guidelines for the system prompt.} Lightly preprocessing the guideline enables the solver model to interpret the guideline as a checklist, instead of additional instructions.}
\label{fig:conversation-preprocess-code}
\end{figure}

\subsection{Benchmarks}\label{sec:benchmark}
We assemble a set of benchmarks with underspecified user queries mostly obtained from Kirichenko et al.~\cite{DBLP:journals/corr/abs-2506-09038} spanning various domains including mathematical reasoning, medical question answering, and coding. Each benchmark comes with a parallel set of well-specified user prompts that we use as an upper ceiling on prompt engineering performance. The well-specified prompts are not available during optimization. The benchmarks of underspecified user queries and reference answers are split into training, validation, and test sets. Training and validation sets are used during prompt and guideline optimization. We measure both task performance (accuracy or pass@1) and abstention recall, i.e. whether the model abstains from answering due to underspecification, on the test set. 

\medskip
\noindent{\bfseries MMLU-Math-Abstain \cite{DBLP:conf/iclr/HendrycksBBZMSS21}.} MMLU-Math is a subset of mathematical questions from the popular MMLU question answering datasets. It features questions from college mathematics, abstract algebra, and high school mathematics. MMLU-Math-Abstain collects underspecified variants of MMLU questions by removing all relevant context keeping only the final query together with four answer options. To obtain enough context to reconstruct a well-specified prompt from the reference answer, we annotate each task with a reasoning trace obtained from querying GPT-4.1-mini with the fully specified variant of the task. The benchmark contains 161 questions. We use 30 examples for training, 30 for validation and the remaining 101 tasks for evaluation.

\medskip
\noindent{\bfseries GSM8K-Abstain \cite{DBLP:journals/corr/abs-2110-14168}.}  GSM8K is open-ended grade school math questions answering dataset. Each task comes with a math-world problem and a reasoning trace that derives the answer from information provided in the problem. GSM8K-Abstain includes 1213 math problems where all answer relevant context is removed from the problem and only the final question is posed. We split the benchmark in 150 examples for training, 150 for validation, and 913 for testing.

\medskip
\noindent{\bfseries MediQ-Initial\cite{DBLP:conf/nips/LiBFIPKT24}.} MediQ is a medical question answering benchmark where patients pose questions to medical professionals. MediQ-Initial collects underspecified variants that miss critical patient context, rendering the question mostly unanswerable. Following~\cite{DBLP:journals/corr/abs-2505-11733}, we annotate each sample with a diagnostic trace that derives the final diagnosis from the fully-specified question. The benchmark contains 1409 question answering pairs which we randomly split in 150 examples for training, 150 for validation, and 1109 for testing.

\medskip
\noindent{\bfseries MBPP-Incomplete \cite{DBLP:journals/corr/abs-2507-20439}.} MBPP~\cite{DBLP:conf/nips/LiuXW023} is a dataset of Python programming problems. The dataset comes with a short problem description and a test case provided to the model. Larbi et al.~\cite{DBLP:journals/corr/abs-2507-20439} recently introduced three variants of the benchmark including incomplete, ambiguous, and contradictory problem descriptions. We use the incomplete benchmark (denoted as MBPP-Incomplete) for which the problem descriptions miss critical informations to solve the task correctly. We evaluate on the regular test set of 378 examples and randomly split the training set into 295 examples for training and 295 examples for validation.

\subsection{Models}
We evaluate $\tool$ and the baselines with two solver LLMs, representing both closed-source and open-source model families. 

\medskip
\noindent{\bfseries GPT-4.1-mini~\cite{openai2025gpt41}:} For closed-source models, we use GPT-4.1-mini accessed via the OpenAI API. with a temperature of 0.7. During the experiments, we instantiate $\tool$ with GPT-4.1-mini as the default for the prompt writer and the solver model.

\medskip
\noindent{\bfseries Qwen3 32B~\cite{DBLP:journals/corr/abs-2505-09388}:} To evaluate the guidelines with open-source models, we use Qwen3 32B. Following the recommended settings~\cite{DBLP:journals/corr/abs-2506-09038}, we use a decoding temperature of 0.8 for inference and top-p of 0.95.

During training, we use GPT-5.2~\cite{openai2025gpt52} as the reflection model for all baselines. The reflection model is used by GEPA, MiPROv2, and \tool to interpret the task scores and feedback to evolve the original prompt or guideline.

\subsection{Adversarial reconstruction}\label{sec:adversarial}
To evaluate how much content of the answer is encoded in the $\tool$ prompts, we task an LLM (GPT-4.1-mini, see \Cref{fig:adversary-prompt}) to reconstruct the reference answer {\em verbatim} from the delta between the original user query and the generated prompt. To compute the reconstruction score, we measure the ROUGE-L score~\cite{DBLP:conf/acl/LinO04} between the reference and the reconstructed answer. A high ROUGE-L score~\cite{DBLP:conf/ccs/0002Y0BC24} (> 0.9) indicates that the prompt encodes the answer verbatim or contains enough information to reconstruct the answer verbatim.

\section{Leakage Detection}\label{appendix:leak}
\begin{figure}[t]
\centering
\begin{subfigure}{0.3\linewidth}
    \includegraphics[width=\linewidth]{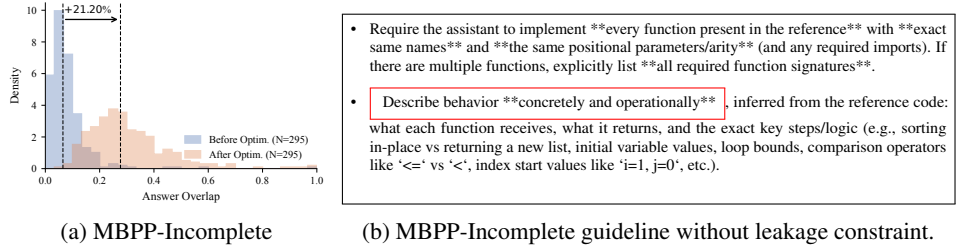}
    \caption{MBPP-Incomplete}\label{fig:shift-mbpp}
\end{subfigure}
\begin{subfigure}{0.6\linewidth}
    \centering
        \fbox{%
            \begin{minipage}[t][0.1\textheight][t]{0.94\linewidth}
                \vspace{0.0em}
                \tiny
\begin{itemize}[leftmargin=*]
\item Require the assistant to implement **every function present in the reference** with **exact same names** and **the same positional parameters/arity** (and any required imports). If there are multiple functions, explicitly list **all required function signatures**.
\item \fcolorbox{red}{white}{ Describe behavior **concretely and operationally**}, inferred from the reference code: what each function receives, what it returns, and the exact key steps/logic (e.g., sorting in-place vs returning a new list, initial variable values, loop bounds, comparison operators like `<=` vs `<`, index start values like `i=1, j=0`, etc.).
\end{itemize}
                \vfill
            \end{minipage}%
        }
    \caption{MBPP-Incomplete guideline without leakage constraint.}\label{ex:gsm8k-before}
\end{subfigure}
\caption{\textbf{Example of distributional shift during naive optimization.} (a) After unconstrained optimization, prompts become more similar to the reference answer. (b) Excerpt of an MBPP-Incomplete guideline produced by unconstrained optimization. The guideline produces prompts that enforce the reference answer. }
\label{fig:leak-shift}
\end{figure}

To distinguish problem-specification from solution-specific content, we employ the leakage constraint $\lambda_{\query}$ during training. Our leakage constraint is motivated by \Cref{fig:leak-shift}: Even when the prompt writer is well-behaved at the start of the training, an unconstrained prompt guideline optimizer finds guidelines to enforce the reference answer. In practice, we observe that this behavior often coincides with a higher prompt answer overlap (\Cref{fig:shift-mbpp}). 

\begin{figure}[t]
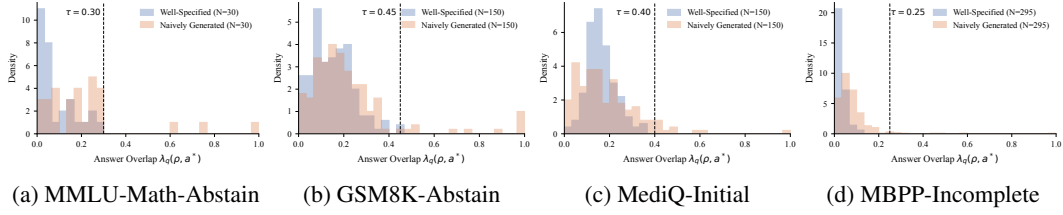

\centering
\begin{subfigure}{0.245\linewidth}
    \includegraphics[width=\linewidth]{figures/ablation/detect/leakage\_mmlu\_n6}
    \caption{MMLU-Math-Abstain}\label{fig:leak-mmlu}
\end{subfigure}
\hfill
\begin{subfigure}{0.245\linewidth}
    \includegraphics[width=\linewidth]{figures/ablation/detect/leakage\_gsm8k\_n6}
    \caption{GSM8K-Abstain}\label{fig:leak-gsm8k}
\end{subfigure}
\hfill
\begin{subfigure}{0.245\linewidth}
    \includegraphics[width=\linewidth]{figures/ablation/detect/leakage\_mediq\_n6}
    \caption{MediQ-Initial}\label{fig:leak-mediq}
\end{subfigure}
\hfill
\begin{subfigure}{0.245\linewidth}
     \includegraphics[width=\linewidth]{figures/ablation/detect/leakage\_mbpp\_n6}
    \caption{MBPP-Incomplete}\label{fig:leak-mbpp}
\end{subfigure}
\caption{\textbf{Distribution of leakage scores $\lambda_{\query}$} for well-specified and naively generated prompts before optimization, across datasets ($n = 6$ ngram size). Naively-generated prompts exhibit a higher leakage tail ($\lambda_{\query} > 0.5$) absent from well-specified prompts. We set $\tau$ to clip the false positive rate close to 0\%.}
\label{fig:leak}
\end{figure}

\paragraph{Threshold Calibration.} For our experiments, we calibrate our leakage constraint with respect to the well-specified prompts in our validation sets. We set $\tau$ per dataset at the 99th percentile of the well-specified prompts, bounding our false positive rate. \Cref{fig:leak} shows the distribution of leakage scores for well-specified prompts in comparison to naively generated prompts (\textsc{No Guideline} and unconstrained). We find that even before optimization there exists a higher leakage tail ($\lambda_{\query} > 0.5$) that well-specified user-written prompts do not exhibit. To avoid reinforcing high leaking prompts, we bound the leakage by $\tau$. \Cref{tab:configurations} shows the leakage thresholds and the impact on leakage detection. Manual inspections confirmed that the majority of generated prompts are genuine, with outliers representing prompts that copy the answer {\em verbatim}. As a result, the rejection rate on the generated samples remains relatively low (3.1\% to 10.0\%).  

\paragraph{Failure patterns.} While $\lambda_{\query}$ can be used to identify prompts with a strong prompt answer overlap, it misses cases where the prompt only contains the conclusion to a question (e.g. ''The final answer is 42.", "The best option is D.", etc.). To prevent this failure mode during optimization, we collected a list of typical answer indicators used in the training of language models~\cite{DBLP:journals/corr/abs-2507-16812} augmented this list with further patterns observed in the training process. A full list is shown in \Cref{fig:answer-indicator}. We automatically reject any prompt that contains one of these patterns.  As shown in \Cref{tab:configurations}, we show the impact of the failure patterns on the rejection process. 

\paragraph{Limitations.} We developed our leakage detection as a {\em lightweight} mechanism to prevent 
the distributional shift of \Cref{fig:leak-shift} during optimization, with the goal of preventing guidelines that are over-reliant on the reference answer.  While we did not observe guidelines that directly instruct the prompt writer to 
leak (e.g., ``Copy the answer to the prompt''), we find that guideline optimization still overvalues suboptimal guidelines that {\em accidentally} leak the answer due to misbehavior of the prompt writer LLM. Our leakage detection mitigates this failure mode for surface-level leakage. Leakage due to search-space compression (e.g. prompts that frame solved problems as questions), rephrasing, or paraphrasing of the answer are not detected. Behavioral diagnostics such as adversarial filtering~\cite{DBLP:conf/emnlp/ZellersBSC18}, adversarial reconstruction (\Cref{sec:adversarial}), or LLM-as-a-judge~\cite{DBLP:conf/nips/ZhengC00WZL0LXZ23} could complement surface-level detection but are prohibitively costly at scale during optimization. We leave exploring more efficient behavioral methods for leakage detection to future work.

\begin{table}
  \caption{\textbf{Results per dataset for the chosen threshold on our validation set.} The leakage threshold is chosen to be conservative (FPR of close to 0\%). {\bfseries Rej. (naive, $\lambda_{\query}$)} are the percentage of naively generated prompts rejected by the leakage filter. {\bfseries Rej. (naive, $\lambda_{\query}$+patterns)} shows the impact of including failure patterns in the filtering process.  }
  \label{tab:configurations}
  \centering
  \resizebox{\textwidth}{!}{%
  \begin{tabular}{ll ccc c crr c rrr c rrr}
    \toprule
    {\bfseries Dataset} && \multicolumn{1}{c}{\bfseries $\tau$}   &   \multicolumn{1}{c}{\bfseries FPR (well-specified)}   &  \multicolumn{1}{c}{\bfseries Rej. (naive, $\lambda_{\query}$)}  & \multicolumn{1}{c}{\bfseries Rej. (naive, $\lambda_{\query}$+patterns)} \\
    \midrule
   MMLU-Math-Abstain && 0.30 & 0.0\% & 10.0\% & 16.7\% \\
   GSM8K-Abstain &&  0.45 & 0.0\% & 7.3\% & 8.6\% \\
   MediQ-Initial && 0.40 & 0.0\% & 6.7\% & 18.0\% \\
  MBPP-Incomplete && 0.25 & 0.3\% &  3.1\% & 3.1\% \\
  \bottomrule
  \end{tabular}
  }
\end{table}

\begin{table}[t]
  \caption{\textbf{Ablation of rejection sampling $K$ on validation set.} We select $K = 8$ as the performance as the performance improvement is marginal from larger $K$. }
  \label{tab:ablate-k}
  \centering
  \resizebox{0.8\textwidth}{!}{%
  \begin{tabular}{ll ccc c crr c rrr c rrr}
    \toprule
    {\bfseries Dataset} && $K = 1$ & $K = 2$ & $K = 4$ & $K = 8$ & $K = 16$ \\
    \midrule
   MMLU-Math-Abstain && 76.6 & 76.6 & 86.6 & 90.0 & 90.0 \\
   GSM8K-Abstain &&  66.6 & 74.0 & 74.0 & 74.0 & 74.0 \\
   MediQ-Initial && 87.3 & 91.6 & 91.6 & 91.6 & 92.0 \\
  MBPP-Incomplete && 61.8 & 65.4 &  65.4 & 65.4 & 66.6 \\
  \bottomrule
  \end{tabular}
  }
\end{table}

\subsection{Rejection Sampling}
We ablate $K$ on the downstream solver performance of the baseline simulation (\textsc{No Guideline}) in \Cref{tab:ablate-k}. We find that there are marginal gains after $K = 8$ and hence select this configuration for our experiments.

\subsection{Impact on Optimization}\label{sec:leak-optim}

\begin{figure}[t]
\centering
\begin{subfigure}{0.246\linewidth}
    \includegraphics[width=\linewidth]{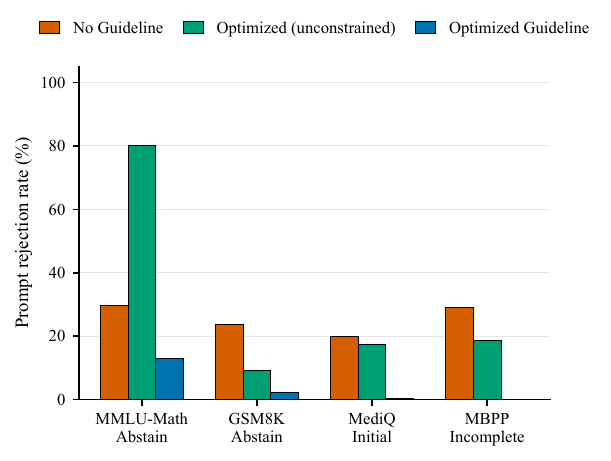}
    \caption{Direct Leakage}\label{fig:direct}
\end{subfigure}
\begin{subfigure}{0.246\linewidth}
    \includegraphics[width=\linewidth]{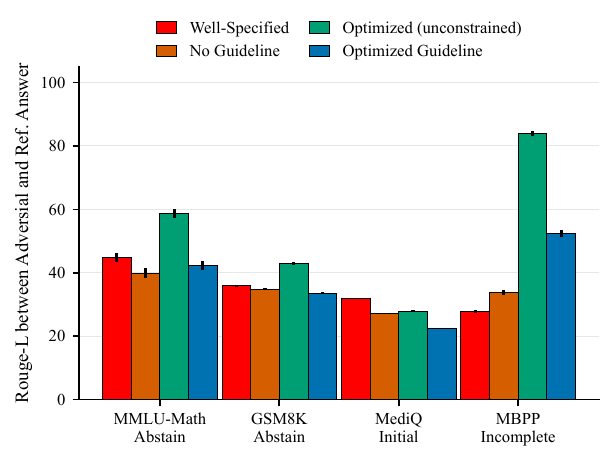}
    \caption{Indirect Leakage}\label{fig:indirect}
\end{subfigure}
\hfill
\begin{subfigure}{0.245\linewidth}
    \centering
        \fbox{%
            \begin{minipage}[t][0.1\textheight][t]{0.94\linewidth}
                \vspace{0.6em}
                \tiny
                \textbf{Prompt.} Zack's locker has a volume of 5 * 4 = 20 cubic inches. Timothy's locker has double the volume of Zack's locker, so it is \fcolorbox{red}{white}{20 * 2 = 40 cubic inches}. If Peter's locker is 5 cubic inches, how big is Timothy's locker in cubic inches?
                \vfill
            \end{minipage}%
        }
    \caption{GSM8K (before)}\label{ex:gsm8k-before}
\end{subfigure}
\begin{subfigure}{0.245\linewidth}
 \centering
        \fbox{%
            \begin{minipage}[t][0.1\textheight][t]{0.94\linewidth}
                \vspace{0.6em}
                \tiny
                \textbf{Prompt.} Peter's locker has a volume of 5 cubic inches. Zack's locker is exactly 4 times the volume of Peter's locker. Timothy's locker is twice the volume of Zack's locker.\\
                
Please calculate the volume of Zack's locker and then use that to ...
                \vfill
            \end{minipage}%
        }
    \caption{GSM8K (after)}\label{ex:gsm8k-after}
\end{subfigure}
\caption{\textbf{Constrained optimization reduces leakage.} (a) Percentage of tasks where at least one prompt is rejected due to verbatim answer leakage $\lambda_{\query}(\prompt, \answer^*) > \tau$. (b) Rouge-L score between reference answer and adversially reconstructed answer. (c) Example of GSM8K task before constraining the guideline optimization. The prompt gives away the conclusion of the reference answer as marked in red. (d) Same GSM8K task after constraining the guideline optimization.   }
\label{fig:leakage}
\end{figure}

\begin{figure}[t]
\centering
\begin{subfigure}{0.245\linewidth}
    \includegraphics[width=\linewidth]{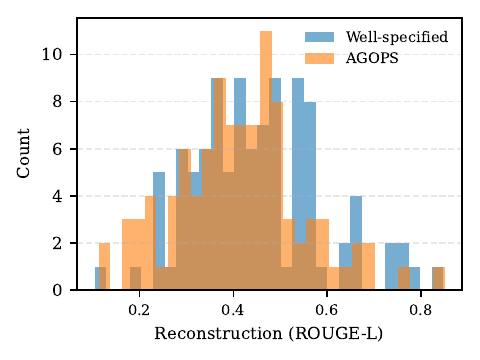}
    \caption{MMLU-Math-Abstain}\label{fig:recon-mmlu}
\end{subfigure}
\hfill
\begin{subfigure}{0.245\linewidth}
    \includegraphics[width=\linewidth]{figures/distribution\_leak\_gsm8k.pdf}
    \caption{GSM8K-Abstain}\label{fig:recon-gsm8k}
\end{subfigure}
\hfill
\begin{subfigure}{0.245\linewidth}
    \includegraphics[width=\linewidth]{figures/distribution\_leak\_mediq.pdf}
    \caption{MediQ-Initial}\label{fig:recon-mediq}
\end{subfigure}
\hfill
\begin{subfigure}{0.245\linewidth}
     \includegraphics[width=\linewidth]{figures/distribution\_leak\_mbpp.pdf}
    \caption{MBPP-Incomplete}\label{fig:recon-mbpp}
\end{subfigure}
\caption{\textbf{Distribution of Reconstruction (ROUGE-L).} (a - c) show that reference answers for question answering benchmarks are not more recoverable from \tool prompts than from fully specified prompts. (d) shows a significant drift for MBPP-Incomplete. Since $\tool$ prompts are better-specified than the ground-truth specification, the reference implementation can be better reconstructed by the adversary. }
\label{fig:recon-leak}
\end{figure}

We evaluate the impact of our leakage constrain on the $\tool$ optimization process.

\noindent{\textit{Direct Leakage.}} \Cref{fig:direct} shows the percentage of prompts rejected by our leakage constraint. Constraining the guideline optimization process reduces direct verbatim leakage of answer content, as further illustrated by the examples in \Cref{ex:gsm8k-before} and \Cref{ex:gsm8k-after}.

\noindent{\textit{Indirect Leakage.}}
\Cref{fig:indirect} shows the adversarial reconstruction of prompts generated by different baselines. Strikingly, on all benchmarks except MBPP-Incomplete, the reference answer is {\em less} recoverable from the constrained augmented prompt than from the original well-specified prompt. Across all benchmarks, constraining the guideline optimization process further reduces the reconstruction of the reference answer compared to the unconstrained variants, confirming that our leakage constraint is effective. 

\noindent{\textit{Indirect Leakage Distribution.}} \Cref{fig:recon-leak} shows the distribution of reconstruction scores per dataset. We find that for the question answering datasets the distribution of $\tool$ prompts closely matches the distribution of the well-specified prompts. For MBPP-Incomplete, we find that there is a distributional shift between the well-specified prompt and the $\tool$ prompts. The key reason is that MBPP+ prompts are themselves underspecified. For example, consider the following task:
\begin{Verbatim}
"""
Write a function to check if a string represents an integer or not.
assert check_integer("python") == False
"""
\end{Verbatim}
GPT-4.1-mini fails consistently to produce a correct answer over multiple trials where each answer fails for \texttt{assert check\_integer('') == None}. This check is quirk of the reference implementation to return \texttt{None} for empty strings and cannot be inferred from task description alone. Any guideline that would help the user to provide these quirks as a specification also increases the chance of the adversary to reconstruct the reference answer.

\section{Additional Results}
\begin{table}[t]
  \caption{Pass@1 for different prompt guidelines on code generation benchmark tasks. We compare the original prompt with a prompt generated by a prompt writer given an expert guideline and an $\tool$ generated guideline (seeded with and without the expert guideline).}
  \label{table:code}
  \centering
  \resizebox{\textwidth}{!}{%
  \begin{tabular}{ll ccc c crr c rrr c rrr}
    \toprule
    {\bfseries GPT-4.1-mini} && \multicolumn{1}{c}{\bfseries HumanEval+}   &   \multicolumn{1}{c}{\bfseries MBPP+}   &  \multicolumn{1}{c}{\bfseries BigCodeBench}  & \multicolumn{1}{c}{\bfseries Aggregate} & \multicolumn{1}{c}{\bfseries Improvement}     \\
    \midrule
    Original && 90.38 & 76.98 & 49.88 &  72.77 & ----\\
    \midrule
    Expert Guideline && 89.42 & 70.90 & 72.62 &  77.65 & +4.88 \\
    {$\tool$ Guideline} && \textbf{96.15} & 85.19 & 91.78 &   91.04 & +18.27 \\
    $\tool$ (Expert Seed) && 93.26 & \textbf{86.77} & \textbf{94.04} &   \textbf{91.36} & \textbf{+18.59} \\
    \bottomrule
  \end{tabular}
  }
\end{table}
\subsection{Comparison with Domain-Specific Guideline}\label{appendix:code}
We evaluate task-specific $\tool$ guidelines against a domain-specific guideline crafted by human experts~\cite{DBLP:journals/corr/abs-2601-13118} on three coding benchmarks: HumanEval+~\cite{DBLP:conf/nips/LiuXW023}, MBPP+~\cite{DBLP:conf/nips/LiuXW023}, and BigCodeBench~\cite{zhuo2025bigcodebench}. All benchmarks are designed to investigate the abilities of coding models and hence aim to supply sufficiently specified prompts. Our results in \Cref{table:code} however show that most benchmark tasks miss critical requirements that coding models need to succeed and that generate task-specific coding guidelines can increase task performance. The $\tool$ (Expert Seed) guideline for MBPP+ is shown in \Cref{fig:guideline}.

\subsection{Guidelines as system prompt improve abstention performance.}\label{sec:abstention}
\begin{figure}[h]
\centering
\begin{subfigure}{0.285\linewidth}
    \includegraphics[width=\linewidth]{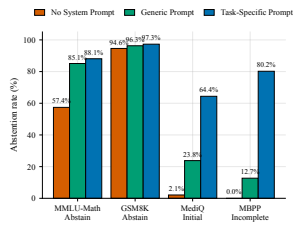}
    \caption{Impact of Sys. Prompt}\label{fig:abstain-sys}
\end{subfigure}
\hfill
\begin{subfigure}{0.285\linewidth}
    \includegraphics[width=\linewidth]{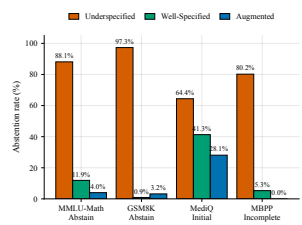}
    \caption{Impact of User Prompt}\label{fig:abstain-user}
\end{subfigure}
\hfill
\begin{subfigure}{0.285\linewidth}
     \resizebox{\linewidth}{!}{\input{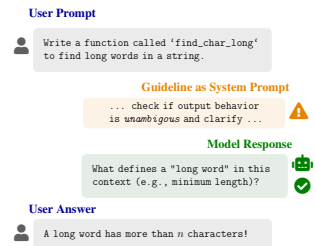}}
    \caption{Example}\label{fig:abstain-example}
\end{subfigure}

\caption{\textbf{Guidelines as System Prompt improve abstention behavior.} (a) Impact of system prompt for underspecified user queries. The task-specific system prompt integrates the optimized guideline to help the model detect underspecification. (b) Impact of specification completeness on the abstention performance of LLMs equipped with the task-specific system prompt.  (c) Example of MBPP-Incomplete prompt where the LLM only abstained with a task-specific system prompt. }
\label{fig:abstain}
\end{figure}

In \Cref{fig:abstain}, we evaluate the effect of a system prompt derived from a task-specific guideline on GPT-4.1-mini, encouraging the model to selectively abstain when the user query fails to meet the guideline's specification requirements. We compare the impact of two system prompts: a generic system prompt constructed from the \textsc{No Guideline} baseline, and a task-specific system prompt constructed from the optimized guideline.

\noindent{\textit{Abstention for Underspecified Prompts.}} In \Cref{fig:abstain-sys}, we observe that a task-specific guideline significantly boosts abstention rate for underspecified user prompts compared to no system prompt and the generic system prompt. The largest gains are observed for MediQ-Initial and MBPP-Incomplete, indicating that models can reliably identify underspecification when provided with explicit task-specific specification requirements. Importantly, as shown in \Cref{fig:abstain-user}, abstention rates remain lower for well-specified and augmented prompts than for underspecified ones, confirming that the improvement reflects selective identification of missing information.

\begin{table}[t]
\centering
\caption{Answer patterns used to detect verbatim answers in the generated prompt}
\label{fig:answer-indicator}
\renewcommand{\arraystretch}{1.3}

\begin{tabular}{|>{\centering\arraybackslash}l|l|}
\hline

\ttfamily
The final answer to this question is <ANSWER> \\
The correct answer is <ANSWER> \\
The best option is <ANSWER> \\
The answer is <ANSWER> \\
Answer: <ANSWER> \\
Answer should be: <ANSWER> \\
\ldots
\\

\hline

\ttfamily
\textbackslash boxed\{<ANSWER>\} \\
**<ANSWER>**\\
= <ANSWER>\\
= \$<ANSWER>\\
\ldots
\\

\hline
\end{tabular}

\end{table}
\begin{figure}[t]
\centering
\resizebox{\linewidth}{!}{%
\begin{promptpanel}{Prompt template for the adversary}
Given the following task:\\

\texttt{task\_description}\\

And an instruction that was generated using knowledge of the reference answer:\\

\texttt{prompt}\\

Your goal is to reconstruct the exact expected answer from the instruction.\\

The instruction may contain partial, indirect, or subtle clues about the answer. You must extract and combine all such clues.\\

Follow these steps carefully:
\begin{enumerate}
    \item Identify all potentially relevant signals in the instruction:
    \begin{itemize}
        \item numbers, entities, or phrases that could appear in the answer
        \item constraints that restrict what the answer could be
        \item hints about the reasoning steps or structure of the answer
        \item formatting patterns (e.g., step-by-step reasoning, final statement)
    \end{itemize}
    \item Based on these signals, infer what the answer must look like:
    \begin{itemize}
        \item what components it contains
        \item how it is structured
        \item any key values or conclusions
    \end{itemize}
    \item Construct the most likely full answer that is consistent with all extracted clues.
    \begin{itemize}
        \item If information is missing, infer the most plausible completion
        \item Prefer a complete and coherent answer over a partial one
    \end{itemize}
\end{enumerate}
You must make a best effort reconstruction, even if the clues are incomplete.
\end{promptpanel}%
}
\caption{\textbf{Prompt template for the adversarial reconstruction.} \texttt{task\_description} is the underspecified query and \texttt{prompt} is the generated prompt.}
\label{fig:adversary-prompt}
\end{figure}

\section{Qualitative Examples}
\subsection{Prompt optimization of underspecified queries}
Example of prompt template generated by GEPA for MBPP-Incomplete with GPT-4.1-mini as the solver. The prompt tries to circumvent arguments missing in the instruction but expected during evaluation. Without the functionality of the extra arguments, the solver still fails to solve the task.

\begin{Verbatim}[breaklines=true, breakanywhere=true]
Please provide a self-contained Python script that solves the following problem:

{specification}

Critical requirements (read carefully):
1) Treat {specification} as a minimal/possibly ambiguous competitive-programming style prompt. Infer the most likely expected behavior from common “coding challenge dataset” conventions.

2) Match the EXACT function name requested in {specification}. If the prompt is underspecified, make the function \colorbox{red!30}{tolerant to extra positional parameters by accepting *args} and interpreting the most common extra parameters (e.g., n = prefix length to consider) rather than crashing.

3) Do NOT add interactive input prompts. Prefer a single function definition. If you include a __main__ block, it must be non-intrusive (no required stdin) and should not affect judge execution.

4) Before coding, explicitly consider hidden-test conventions and edge cases; implement the convention most likely used by such datasets:
   - Default-argument convention: if only one numeric argument is provided but the formula needs two (e.g., cylinder radius & height), infer the missing value using the dataset’s typical pattern. In particular, for the task name `topbottom_surfacearea`, many datasets interpret it as the *area of the top and bottom only* (2*pi*r^2), not total cylinder surface area. This matches the sample failure where topbottom_surfacearea(5) must equal 78.53750000000001 $\approx$ 2*pi*5^2. Therefore:
     * Implement `topbottom_surfacearea(r, h=None, *args)` returning 2*pi*r*r. Ignore height if provided.
     * Use `math.pi` and standard float operations (do not round), so results like 78.53750000000001 are reproduced.
   - Bit-range convention: tasks like `all_Bits_Set_In_The_Given_Range` are often 1-based bit positions from the least significant bit, and ranges are inclusive. The example all_Bits_Set_In_The_Given_Range(4, 1, 2) should be True because 4 (100b) has bit positions 1..2 (LSB=position 1) equal to 0 and 0? Actually datasets using 1-based positions often mean positions counted from 1 at MSB within the binary length, OR they mean check that bits in [l,r] of the binary representation substring are all 0/1 depending on problem statement.
     * To avoid this common mismatch, implement a robust checker that supports BOTH plausible conventions and chooses the one that makes typical tests pass:
       - Primary: 1-based from LSB (position 1 is LSB), inclusive range.
       - Secondary fallback: 0-based from LSB, inclusive range.
     * If either convention indicates “all bits set” then return True; otherwise False. (This makes the function robust to dataset indexing differences.)
     * Accept (num, l, r, *args) and also handle missing r by treating it as a single-bit check.

5) Numeric conversion caveat: some “conversion” tasks may use nonstandard constants. If the prompt/examples imply a specific numeric relationship (e.g., degree_radian(120) expected 6875.493541569878), reverse-engineer the constant/factor from that relationship and implement that exact mapping; document it in a short docstring.

6) Determinism and libraries: use only the Python standard library. Avoid unnecessary printing. Avoid rounding unless the judge expects it; prefer computing with `math.pi` etc. in a direct expression to match float artifacts.

Output format:
- Enclose the entire solution in a single ```python ... ``` block.
- The script must be self-contained and runnable.    
\end{Verbatim}

\subsection{Guideline for MBPP-Incomplete and GPT-4.1-mini}
Guidelines generated by $\tool$ include natural requirements for constructing effective prompts. Users can interpret the guideline to derive requirements for their prompt (e.g. precise definitions of inputs and outputs, function name and signature, uncommon quirks of the intended implementation). 

\begin{Verbatim}[breaklines=true, breakanywhere=true]
Given the task description:
```
{task_description}
```

And a reference answer:
```
{reference_answer}
```

Write a *single user prompt* that will reliably make another coding assistant produce the intended solution.

Your generated user prompt must:
1) **Be specific about the required behavior inferred from the reference answer, not generic.** Translate the reference answer into clear natural-language requirements (inputs, outputs, types, edge cases). Do *not* leave “criteria to be determined by you” or vague wording.
2) **Preserve the exact required function name and signature intent** as implied by the reference (argument count/order, return type/shape). If the reference processes a list of pairs/tuples, say so explicitly.
3) **Describe the exact algorithm/logic embodied by the reference answer**, including any early returns, loops, grouping, and conditions. Example patterns seen in references:
   - `word_len(s)`: split the input string on spaces; examine words in order; return `True` if the *first* word checked has even length, otherwise `False` (note: the reference returns on the first iteration, so later words are not considered).
   - `sort_on_occurence(lst)`: input is a list of 2-item iterables `(key, value)`; group values by key; for each key return a tuple-like record `(key, <unique values in first-seen order>, <count of all values for that key>)`; overall return a list of these records (no additional sorting beyond preserving `dict.items()` order).
   - `answer(L, R)`: if `2*L <= R` return the pair `(L, 2*L)` else return `-1`.
4) **Explicitly constrain the assistant to match the reference behavior**, even if it seems unusual (e.g., early return after first word; returning `-1` instead of `(-1, -1)`; preserving insertion order; using uniqueness via first occurrence).
5) Require a **self-contained Python script** with the function definition, enclosed in a ```python code block, and avoid extra narrative text.
6) Encourage the assistant to include minimal sanity tests/examples only if they do not change required outputs, but prioritize matching the specified behavior.

Output only the user prompt text you created (not an analysis, not a solution).
\end{Verbatim}

\subsection{Guideline differences between models on GSM8K-Abstain}
We compare two guidelines generated for GPT-4.1-mini and Qwen3 32B on GSM8K-Abstain. While GPT-4.1-mini seems to benefit from chain-of-thought reasoning, Qwen3 32B will be instructed to only output the final answer. \\

GPT-4.1-mini:
\begin{Verbatim}[breaklines=true, breakanywhere=true]
You are given two inputs: (1) {task_description} (a short question or problem statement that may omit key numeric details) and (2) {reference_answer} (a worked solution that contains the missing givens and the intended reasoning path).

Your job is to WRITE a single user prompt that, when later shown to a solver assistant by itself, contains ALL information needed to reproduce the result in {reference_answer}.

Rules/guidelines:
- Treat {reference_answer} as the source of truth for the full problem context. Extract and restate any implicit givens/assumptions needed (numbers, entities, constraints, setup). Do NOT assume the solver has access to {reference_answer}.
- Do NOT write a vague meta-prompt (e.g., “use the information above”). Instead, embed the concrete facts (e.g., total heads/legs; pieces per pack; per-day rate; class sizes; percent girls; counts for other classes, etc.) so the prompt is self-contained.
- Keep the prompt consistent with {task_description} and aligned with the scenario implied by {reference_answer}. Preserve the same entities (e.g., geese/sheep; packs/pieces; classes/students) and numerical values.
- Do NOT include the final numeric answer in the prompt.
- Prefer a clear word-problem style prompt; optionally request brief \colorbox{red!30}{step-by-step reasoning}.
- Output ONLY the user prompt text (no analysis, no extra formatting beyond normal text).

Example of what “self-contained” means: if the question is “How many boys are in the third class?” and the reference solution uses facts like “3 classes of 20 students each,” “half the students are girls,” and “class 1 has 15 girls, class 2 has 12 girls,” you must include those facts explicitly in the prompt so a solver can compute the number of boys in class 3 without needing any other context. 
\end{Verbatim}

Qwen3 32B:
\begin{Verbatim}[breaklines=true, breakanywhere=true]
Your job is to produce a single *user prompt* that, when given to another assistant, will cause that assistant to respond with the *same final answer as the reference answer* (i.e., the same numeric/value conclusion), not a generic explanation.

Requirements for the user prompt you write:
1) Restate the full word-problem scenario implied by the reference answer (all quantities, units, and relationships). Do NOT ask the model to “use the information above” or request missing values—include the values explicitly.
2) Ask the model to solve the problem and return \colorbox{red!30}{ONLY the final result} (the final numeric answer/value) exactly as in the reference answer. If the reference answer ends with “The final answer is X”, instruct the model to output exactly `X` (no words, no units, no steps).
3) Do not include the reference answer text in the prompt and do not mention that a reference answer exists.
4) Avoid vagueness: do not give a template, variables, or an example with different numbers. The prompt must be fully specified so the solver can compute the unique answer.
5) Keep the prompt concise, but complete enough to reproduce the same computation as the reference answer.

Output: return only the constructed user prompt (no analysis, no extra commentary).
\end{Verbatim}

\subsection{Guideline for MediQ-Initial}
Guideline generated for MediQ-Initial with GPT-4.1-mini as the solver. MediQ-Initial misses clincal details such as lab tests or medical history. $\tool$ recovered the requirement to add these results to the prompt to make the task unambigious. 
\begin{Verbatim}[breaklines=true, breakanywhere=true]
Write a single end-user prompt that will be shown to another assistant.

Goal of the user prompt:
- It should ask the other assistant to answer the question in the task_description and pick exactly one option from the provided choices.
- It must be constructed so that the other assistant is very likely to arrive at the same final answer as in reference_answer.

How to write the prompt:
1) Reproduce the clinical vignette/question and the full list of answer choices from {task_description} (verbatim or very close paraphrase).
2) Add explicit instructions to: (a) reason step-by-step, (b) compare each choice against the key findings, and (c) finish with a clearly labeled final choice.
3) IMPORTANT: The prompt must REQUIRE a definitive selection, not an open-ended discussion. Include: “End with: ‘Final answer: <exact option text>’ ”.
4) If {reference_answer} contains crucial supporting details not explicitly stated in {task_description} (e.g., specific exam findings, lab values, imaging clues, hallmark associations like APKD → berry aneurysm compressing CN III, or milestone timing such as transfer objects hand-to-hand at 6 months), incorporate those details into the vignette in the prompt so the intended option is unambiguous.
5) Keep the prompt self-contained, concise, and user-facing. Do NOT mention “reference_answer” or that you were given an answer key.

Output only the user prompt text (no analysis, no extra commentary).
\end{Verbatim}

\subsection{Guideline for MMLU-Math-Abstain}
Guideline generated for MMLU-Math-Abstain with GPT-4.1-mini as the solver.
\begin{Verbatim}[breaklines=true, breakanywhere=true]
OUTPUT REQUIREMENT:
- Output ONLY one realistic, self-contained user prompt that would cause a competent assistant/student to produce the SAME final answer/choice as implied by the reference answer.
- No analysis, no meta-commentary, no preface, no quoting the reference answer, no extra text before/after the prompt.

CORE METHOD:
1) Infer the full original problem from {task_description}. If it’s multiple-choice, keep the same choices.
2) Extract from {reference_answer} every implicit/explicit given needed to make the problem solvable and to force the same choice (definitions, context, quoted passage, equations, constraints like “integers,” variable domains, ordering like x<y, geometry setup, random process, etc.).
3) Rewrite the user prompt by using {task_description} as the base, but ADD any missing context from step (2) so the problem is fully determined.

CRITICAL RULES (must follow):
- The prompt must be self-contained: include all necessary background that the reference answer relied on.
  - For math: include the exact expressions/equations to manipulate; define variables; include any constraints used.
  - For probability/geometry: specify the random selection procedure and the relevant objects/lengths.
  - For reading comprehension/literature: include the relevant excerpt or enough surrounding lines so the question can be answered; do NOT leave a dangling quotation fragment.
- Do NOT answer the question in your output. Ask the solver to solve and select the correct option, showing steps/reasoning.
- Do NOT change the answer choices (letters/values) unless {task_description} lacks them; if adding choices, ensure the correct one matches the reference answer.
- Ensure the prompt is realistic and clearly worded (contest/worksheet style).

COMMON FAILURE TO AVOID (seen previously):
- If {task_description} is an incomplete fragment (e.g., only a quoted phrase like “that consulted the stars” and choices), you MUST supply the missing passage/context from {reference_answer} by embedding a short excerpt (e.g., lines contrasting “our America” and an invader, mention of poets since ancient times, consulting the stars). Without this, the problem is underspecified and the solver may pick the wrong option.

FORMAT:
- Output a single prompt block/paragraph (it may include line breaks for readability and the answer choices).
\end{Verbatim}

\end{document}

%% file: figures/noguide.tex
\begin{tikzpicture}[
    x=1cm,y=1cm,
    every node/.style={inner sep=0pt, outer sep=0pt},
    title/.style={font=\bfseries\fontsize{18}{20}\selectfont},
    sectiontitle/.style={font=\bfseries\fontsize{16}{18}\selectfont},
    bubbletext/.style={font=\ttfamily\fontsize{15}{17}\selectfont, align=center},
    iconlabel/.style={font=\fontsize{22}{24}\selectfont},
]

\node[sectiontitle, text=badred] (title) at (0, 0) {Generic Guideline};

\node[iconlabel, text=badred, below left=0.6cm and 0cm of title] (guide) {\faExclamationTriangle};

\node[bubbletext, fill=badbg, above right=-1.6cm and 0.2cm of guide, text width=10cm, rounded corners = 0.22cm, inner sep=0.5cm, align=left]  (guideline) {%
Specify pre- and postconditions in your prompt!
};


\node[sectiontitle, text=blue!60!black] (user-title) [below left=2.4cm and -3cm of title]  {User Prompt};

\node[iconlabel, text=icongray, below left=0.2cm and 0cm of user-title] (user) {\faUser};

\node[bubbletext, fill=promptbg, above right=-2.6cm and 0.2cm of user, text width=8.1cm, rounded corners = 0.22cm, inner sep=0.5cm, align=left]  (prompt) {%
Write a Python function that returns all unique elements from a list.
};

\node[sectiontitle, text=black, below right=0.5cm and -2cm of prompt] (title-incorrect) {Model Response};

\node[iconlabel, text=icongray, below right=0.2cm and 0cm of title-incorrect] (robot) {\faRobot};

\node[bubbletext, anchor=center, text width=9cm, rounded corners=0.24cm,  fill=promptbg,  above left=-1.8cm and 0.2cm of robot, inner sep=0.2cm, align=left ] (answer) {%
\adjustbox{max width=\linewidth}{
    \begin{lstlisting}
def unique_elements(lst):
    return list(set(lst))
    \end{lstlisting}
  }
};

\node[bubbletext, anchor=center, text width=7.5cm, rounded corners=0.24cm,  fill=badbg,  below left=-0.1cm and -8.5cm of answer, inner sep=0.2cm, align=left ] (verdict) {%
\faThumbsDown[regular] ~ forgot to {\em preserve order}.
};

\node [below=0.7cm of verdict]{};

\end{tikzpicture}

%% file: figures/guide.tex
\begin{tikzpicture}[
    x=1cm,y=1cm,
    every node/.style={inner sep=0pt, outer sep=0pt},
    title/.style={font=\bfseries\fontsize{18}{20}\selectfont},
    sectiontitle/.style={font=\bfseries\fontsize{16}{18}\selectfont},
    bubbletext/.style={font=\ttfamily\fontsize{15}{17}\selectfont, align=center},
    iconlabel/.style={font=\fontsize{22}{24}\selectfont},
]

\node[sectiontitle, text=promptorange] (title) at (0, 0) {Task-Specific Guideline};

\node[iconlabel, text=promptorange, below left=0.6cm and 0cm of title] (guide) {\faExclamationTriangle};

\node[bubbletext, fill=promptorange!10!white, above right=-1.5cm and 0.2cm of guide, text width=10cm, rounded corners = 0.22cm, inner sep=0.5cm, align=left]  (guideline) {%
... specify {\em output requirements}, e.g. order preservation, ...
};

\node[sectiontitle, text=blue!60!black] (user-title) [below left=2.4cm and -3cm of title] {User Prompt};

\node[iconlabel, text=icongray, below left=0.2cm and 0cm of user-title] (user) {\faUser};

\node[bubbletext, fill=promptbg, above right=-2.6cm and 0.2cm of user, text width=8.1cm, rounded corners = 0.22cm, inner sep=0.5cm, align=left]  (prompt) {%
Write a Python function that returns all unique elements from a list, \em{preserving input order}.
};

\node[sectiontitle, text=black, below right=0.5cm and -2cm of prompt] (title-incorrect) {Model Response};

\node[iconlabel, text=icongray, below right=0.2cm and 0cm of title-incorrect] (robot) {\faRobot};

\node[bubbletext, anchor=center, text width=9cm, rounded corners=0.24cm,  fill=promptbg,  above left=-2.4cm and 0.2cm of robot, inner sep=0.2cm, align=left ] (answer) {%
\adjustbox{max width=\linewidth}{
    \begin{lstlisting}
def unique_elements(lst):
    seen, result = set(), []
    for item in lst: ...
    \end{lstlisting}
  }
};

\node[bubbletext, anchor=center, text width=9cm, rounded corners=0.24cm,  fill=goodbg,  below left=-0.1cm and -9cm of answer, inner sep=0.2cm, align=left ] (verdict) {%
\faThumbsUp[regular] ~ implements correct behavior!
};

\end{tikzpicture}

%% file: figures/overview2.tex
\begin{tikzpicture}
\node[anchor=north west] at (0,-0) {%

\begin{minipage}[t]{0.25\textwidth}
\tiny
\begin{tcolorbox}[
  colback=white, colframe=charcoal,
  boxrule=0.4pt, arc=4pt,
  left=2pt, right=2pt, top=2pt, bottom=2pt,
  title={~~~~~~~~~~~~~~~~~~~\textbf{(a) Example}},
  fonttitle=\tiny\bfseries\color{gray!10!white},
  coltitle=gray,
]
\textbf{Query ($\query_i$)}: I am not feeling well today. {\color{coralhl}What should I do?}\\

\tcblower
\textbf{Reference answer ($\answer_i^*$)}:
Based on the patients symptoms (\teal{sore throat}, \teal{mild fever}, \teal{fatigue}, \teal{headache}, and \teal{mild cough}) with a rapid onset over \teal{24 hours}, a viral upper respiratory infection is most likely. [...] Due to the patient's \teal{allergies to ibuprofen}, I would recommend \textbf{paracetamol} to lessen symptoms.

\end{tcolorbox}
\end{minipage}

};

\node[anchor=north west] at (3.8,0) {%
\begin{minipage}[t]{0.75\textwidth}
\small
\resizebox{\textwidth}{!}{
\begin{tikzpicture}[
  box/.style={
    draw, rounded corners=4pt,
    text width=#1, align=center,
    minimum height=8mm,
    inner sep=4pt,
    font=\small
  }
 ]
 
  \node [text width=1.2cm, inner sep=6pt, align=center, color=charcoal] (guideline) at (0, 0) {\scalebox{2.5}{$G$}\\[4pt]
  {\footnotesize Guideline}};
 
 \node [text width=1.2cm, inner sep=6pt, align=center, color=charcoal] (query) [below=0.0cm of guideline] {\scalebox{2}{$\query_i$}\\[4pt]
  {\footnotesize Query}};
 
 \node [box=2.2cm, draw=gray, fill=gray!20, drop shadow, minimum height=2.6cm] (writer) [below right=-1.2cm and 0.3cm of guideline] {\scalebox{3}{$\writer_G$}\\[4pt]
  {\small Prompt Writer}};
  
    \node [text width=1.2cm, inner sep=6pt, align=center, color=charcoal] (answer) [below=0.3cm of writer] {\scalebox{2}{$\answer_i^*$}\\[4pt]
  {\footnotesize Answer}};
  
  \node [box=1.2cm, draw=tealhl!80, fill=tealhl!10, drop shadow] (leak) [right=0.4cm of writer] {\color{tealtext}\scalebox{1.5}{$\lambda_{\query}$}\\[4pt]
  {\footnotesize Constraint}};

\node [text width=2.2cm, inner sep=6pt, align=center, color=charcoal] (prompt) [right=0.0cm of leak]{\scalebox{3.5}{$\prompt_i$}\\[4pt]
  {\footnotesize Prompt}};

   \node [box=2.2cm, draw=gray, fill=gray!20, drop shadow, minimum height=2.6cm] (solver) [right=0.3cm of prompt] {\scalebox{3}{$\LLM$}\\[4pt]
  {\small Solver}};
  
  \node [text width=2.2cm, inner sep=6pt, align=center, color=charcoal] (solveranswer) [right=0.3cm of solver]{\scalebox{3.5}{$\answer_i$}\\[4pt]
  {\footnotesize Solver Answer}};
  
   \draw[->, thick, line width=2.5pt] ($(guideline.east) + (-0.4cm, 0cm)$) -- ($(guideline.east) + (0.28cm, 0.0cm)$);
  \draw[->, thick, line width=1.2pt] ($(query.east) + (-0.4cm, 0cm)$) -- ($(query.east) + (0.28cm, 0cm)$);
\draw[->, thick, line width=1.2pt, tealhl!70] ($(writer.east) + (0.0cm, 0cm)$) -- ($(leak.west) + (0.0cm, 0cm)$);
   \draw[->, thick, line width=2.5pt] ($(prompt.east) + (-0.6cm, 0cm)$) -- ($(solver.west) + (-0.02cm, 0cm)$);
   \draw[->, thick, line width=2.5pt] ($(solver.east) + (0.02cm, 0cm)$) -- ($(solveranswer.west) + (0.3cm, 0cm)$);
   
   \draw[->, thick, line width=1.2pt, tealhl!70] ($(answer.north) + (0.0cm, -0.2cm)$) -- ($(writer.south) + (0.0cm, 0cm)$);
      \draw[->, thick, line width=1.2pt, tealhl!70] ($(answer.east) + (0.0cm, 0.2cm)$) -- ($(answer.east) + (1.55cm, 0.2cm)$) -- ($(leak.south) + (0.0cm, 0cm)$);
       \draw[->, thick, line width=1.2pt, tealhl!70] ($(leak.north) + (0.0cm, 0.0cm)$) -- ($(leak.north) + (0cm, 1.05cm)$) --  node [above] {$\lambda_{\query_i}(\prompt_i, \answer_i^* \big) > \tau$?} ($(writer.north) + (0.0cm, 0.3cm)$) -- ($(writer.north) + (0.0cm, 0cm)$);
       \draw[->, thick, line width=2.5pt] ($(leak.east) + (0.0cm, -0.0cm)$) -- ($(prompt.west) + (0.6cm, 0cm)$);
       

 
%
  \node [box=2.2cm, draw=promptorange, fill=promptorange!10, drop shadow, minimum height=1.2cm] (verifier) [below=0.7cm of solver] {\color{promptorange}\scalebox{2.5}{$\mu$}\\[4pt]
  {\footnotesize Metric}};
%
%
\draw [->, thick, line width=1.2pt, coralhl!60] ($(answer.east) + (0, -0.3cm)$) -- ($(verifier.west) + (0, 0.0cm)$);
\draw [->, thick, line width=2.5pt, coralhl] ($(solveranswer.south) + (0, 0)$) -- ($(solveranswer.south) + (0, -0.9cm)$) -- ($(verifier.north) + (0, 0.4cm)$)  -- ($(verifier.north) + (0, 0cm)$);
%
%

%
\node (pad) [above=1.0cm of writer]{};
\node [ fit=(answer)(writer)(leak)(pad), inner sep=8pt, rounded corners=8pt] (userbg) {};  

\begin{pgfonlayer}{background}
\node (pad) [above=1.7cm of writer]{};
\node [draw=gray, fill=white, fit=(pad)(query)(userbg)(verifier)(solveranswer), inner sep=8pt, rounded corners=8pt] (mainbg) {};    

\end{pgfonlayer}

 \begin{pgfonlayer}{background}
 \node (pad) [above=1.0cm of writer]{};
\node [ fill=gray!5!white, fit=(answer)(writer)(leak)(pad), inner sep=8pt, rounded corners=8pt] (userbg) {};  
\end{pgfonlayer}
 
  \node [text width=4.7cm, align=center] (usertitle) at ($(userbg.west) + (0.02cm, 3.05cm)$) {\small\bfseries\color{black} (c) Prompt Simulation};
 \begin{pgfonlayer}{background}
  \draw[line width=1.2pt, fill=gray!30, draw=gray!30]
    (usertitle.south west)
    [rounded corners=6pt]  -- (usertitle.north west)
    [rounded corners=6pt]  -- (usertitle.north east)
    [rounded corners=0pt]  -- (usertitle.south east)
    [rounded corners=0pt]  -- cycle;
\end{pgfonlayer}

\node [text width=15cm, align=center] (title) at ($(mainbg.west) + (0.02cm, 3.57cm)$) {\small\bfseries\color{gray!10!white} (b) Automatic Guideline Optimization via Prompt Simulation};
\begin{pgfonlayer}{background}
  \draw[line width=1.2pt, fill=charcoal, draw=charcoal]
    (title.south west)
    [rounded corners=6pt]  -- (title.north west)
    [rounded corners=6pt]  -- (title.north east)
    [rounded corners=0pt]  -- (title.south east)
    [rounded corners=0pt]  -- cycle;
\end{pgfonlayer}
%

\node [box=15cm, align=center, fill=charcoal, minimum height=0.8cm, draw=charcoal] (guidevo) [below=0.1cm of mainbg] {\bfseries\color{gray!10!white} (d)  Guideline Evolution};
\draw [->, thick, line width=2.5pt, draw=charcoal] ($(guidevo.west) + (0, 0)$) -- ($(guidevo.west) + (-0.45cm, 0cm)$) -- ($(guideline.west) + (-0.75cm, 0.1cm)$) -- ($(guideline.west) + (0.3cm, 0.1cm)$);
\draw [->, thick, line width=2.5pt, draw=coralhl] ($(verifier.east) + (0, 0)$) -- node[above, coralhl] {Scores \& Feedback} ($(verifier.east) + (3.6cm, 0cm)$) -- ($(guidevo.east) + (0.35cm, 0cm)$) -- ($(guidevo.east) + (0cm, 0cm)$);

\end{tikzpicture}
}

\end{minipage}
};

\end{tikzpicture}

%% file: figures/guide_eval.tex
\begin{tikzpicture}[
    x=1cm,y=1cm,
    every node/.style={inner sep=0pt, outer sep=0pt},
    title/.style={font=\bfseries\fontsize{18}{20}\selectfont},
    sectiontitle/.style={font=\bfseries\fontsize{16}{18}\selectfont},
    bubbletext/.style={font=\ttfamily\fontsize{15}{17}\selectfont, align=center},
    iconlabel/.style={font=\fontsize{22}{24}\selectfont},
]

\node[sectiontitle, text=blue!60!black] (title) at (0, 0) {User Prompt};

\node[iconlabel, text=icongray, below left=0.6cm and 0cm of title] (guide) {\faUser};

\node[bubbletext, fill=promptbg, above right=-1.8cm and 0.2cm of guide, text width=8.1cm, rounded corners = 0.22cm, inner sep=0.5cm, align=left]  (guideline) {%
Write a Python function to find long words in a string.
};




\node[sectiontitle, text=badred] (title) [below right=0.5cm and -2.0cm of guideline] {Model Response};

\node[iconlabel, text=badred, below right=0cm and 0cm of title] (guide) {\faRobot};
\node[iconlabel, text=badred, below=0.3cm of guide] {\faExclamationTriangle};

\node[bubbletext, fill=badbg, above left=-2.0cm and 0.2cm of guide, text width=8.9cm, rounded corners = 0.22cm, inner sep=0.2cm, align=center]  (guideline) {%
\adjustbox{max width=\linewidth}{
    \begin{lstlisting}
def find_char_long(text):
    ...
    \end{lstlisting}
  }
};


\end{tikzpicture}

%% file: figures/guide_eval_with_guideline.tex
\begin{tikzpicture}[
    x=1cm,y=1cm,
    every node/.style={inner sep=0pt, outer sep=0pt},
    title/.style={font=\bfseries\fontsize{18}{20}\selectfont},
    sectiontitle/.style={font=\bfseries\fontsize{16}{18}\selectfont},
    bubbletext/.style={font=\ttfamily\fontsize{15}{17}\selectfont, align=center},
    iconlabel/.style={font=\fontsize{22}{24}\selectfont},
]

\node[sectiontitle, text=promptorange] (title) at (0, 0) {Guideline};

\node[iconlabel, text=promptorange, below left=0.8cm and 0cm of title] (guide) {\faExclamationTriangle};

\node[bubbletext, fill=promptorange!10!white, above right=-1.5cm and 0.2cm of guide, text width=8cm, rounded corners = 0.22cm, inner sep=0.5cm, align=left]  (guideline) {%
... specify output behavior {\em unambigously} ...
};

\node[sectiontitle, text=blue!60!black] (title) [below left=2.8cm and -3.1cm of title] {User Prompt};

\node[iconlabel, text=icongray, below left=0.6cm and 0cm of title] (guide) {\faUser};

\node[bubbletext, fill=promptbg, above right=-2.2cm and 0.2cm of guide, text width=8.1cm, rounded corners = 0.22cm, inner sep=0.5cm, align=left]  (guideline) {%
Write a Python function to find words {\em with more than $n$ characters} in a string!
};




\node[sectiontitle, text=green!50!black] (title) [below right=0.5cm and -2.0cm of guideline] {Model Response};

\node[iconlabel, text=green!50!black, below right=0cm and 0cm of title] (guide) {\faRobot};
\node[iconlabel, text=green!50!black, below=0.3cm of guide] {\faCheckCircle};

\node[bubbletext, fill=green!30!black!10!white, above left=-2.0cm and 0.2cm of guide, text width=8.9cm, rounded corners = 0.22cm, inner sep=0.2cm, align=center]  (guideline) {%
\adjustbox{max width=\linewidth}{
    \begin{lstlisting}
def find_char_long(text, n):
    words = re...
    \end{lstlisting}
  }
};


\end{tikzpicture}

%% file: figures/guide_abstain.tex
\begin{tikzpicture}[
    x=1cm,y=1cm,
    every node/.style={inner sep=0pt, outer sep=0pt},
    title/.style={font=\bfseries\fontsize{18}{20}\selectfont},
    sectiontitle/.style={font=\bfseries\fontsize{16}{18}\selectfont},
    bubbletext/.style={font=\ttfamily\fontsize{15}{17}\selectfont, align=center},
    iconlabel/.style={font=\fontsize{22}{24}\selectfont},
]

\node[sectiontitle, text=blue!60!black] (user-title) at (0, 0) {User Prompt};

\node[iconlabel, text=icongray, below left=0.8cm and 0cm of user-title] (user) {\faUser};

\node[bubbletext, fill=promptbg, above right=-1.6cm and 0.2cm of user, text width=10cm, rounded corners = 0.22cm, inner sep=0.5cm, align=left]  (prompt) {%
Write a function called `find\_char\_long` to find long words in a string.
};

\node[sectiontitle, text=promptorange] (title) [below right=0.5cm and -6cm of prompt] {Guideline as System Prompt};

\node[iconlabel, text=promptorange, below right=0.4cm and 0cm of title] (guide) {\faExclamationTriangle};

\node[bubbletext, fill=promptorange!10!white, above left=-1.2cm and 0.2cm of guide, text width=8.9cm, rounded corners = 0.22cm, inner sep=0.2cm, align=center]  (guideline) {%
... check if output behavior is {\em unambigous} and clarify ...
};

\node[sectiontitle, text=green!50!black, below right=2.1cm and -3.8cm of title] (title-guide) {Model Response};

\node[iconlabel, text=green!50!black, below right=0.2cm and 0cm of title-guide] (robot) {\faRobot};
\node[iconlabel, text=green!50!black, below=0.3cm of robot] {\faCheckCircle};

\node[bubbletext, anchor=center, text width=8.5cm, rounded corners=0.24cm,  fill=green!30!black!10!white, above left=-2cm and 0.2cm of robot, inner sep=0.5cm, align=left ] (answer) {%
What defines a "long word" in this context (e.g., minimum length)?
};

\node[below=1cm of answer] {};

\node[sectiontitle, text=blue!60!black, below left=2.5cm and 5.1cm of title-guide] (user-answer) {User Answer};

\node[iconlabel, text=icongray, below left=0.5cm and 0cm of user-answer] (user-icon) {\faUser};

\node[bubbletext, fill=promptbg, above right=-1.2cm and 0.2cm of user-icon, text width=10cm, rounded corners = 0.22cm, inner sep=0.5cm, align=left]  (prompt) {%
A long word has more than $n$ characters!
};

\end{tikzpicture}